%% file: acl_latex.tex
\pdfoutput=1
\documentclass[11pt]{article}

\PassOptionsToPackage{table}{xcolor} %
\usepackage[preprint]{acl}

\usepackage{times}
\usepackage{latexsym}

\usepackage[T1]{fontenc}

\usepackage[utf8]{inputenc}

\usepackage{microtype}
\usepackage{amsmath}

\usepackage{inconsolata}

\usepackage{graphicx}

\usepackage{hyperref}       %
\usepackage{url}            %
\usepackage{booktabs}       %
\usepackage{amsfonts}       %
\usepackage{nicefrac}       %
\usepackage{microtype}      %
\usepackage{xcolor}         %
\usepackage{multirow}
\usepackage{xspace}
\usepackage{graphicx}
\usepackage{comment}
\usepackage{siunitx}
\usepackage{tcolorbox}
\tcbuselibrary{listingsutf8, breakable, skins}
\usepackage{subcaption}

\usepackage{longtable}
\usepackage{booktabs}
\usepackage{array}
\usepackage{colortbl}
\usepackage[export]{adjustbox}
\usepackage{pgfplots}
\usetikzlibrary{patterns}
\pgfplotsset{compat=1.18}

\newcommand{\datasetname}{\textsc{knows}\xspace}
\newcommand{\datasetnamefull}{\textbf{K}nowledge \textbf{N}avigation and \textbf{O}rganized \textbf{W}eb \textbf{S}ynthesis\xspace}

\newcommand{\numtotaltasks}{\textsc{110}\xspace}

\newcommand\sect[1]{\S\ref{#1}}

\title{The Hard Part Comes After Search: Benchmarking Web Agents on Synthesizing, Organizing, and Displaying Knowledge}

\author{
\textbf{Alexander Gill} \quad \textbf{Md Farhan Ishmam} \quad \textbf{Xuyen Nguyen} \quad  \textbf{Neha Bhat} \\
\textbf{Parker Henry DeYoung} \quad \textbf{Fateme Hashemi Chaleshtori} \quad \textbf{Nathan Stringham}\\
\textbf{Kenneth Marino}\thanks{Equal contribution as last authors.} \quad \textbf{Ana Marasovi\'c}\footnotemark[1] \\[4pt]
University of Utah \\
\textbf{\small Correspondence:} \texttt{\small \{alex.gill, farhan.ishmam, kenneth.marino, ana.marasovic\}@utah.edu}
}

\begin{document}
\maketitle
\begin{abstract}
Existing computer-use agent benchmarks do not fully evaluate agents acting as assistants. A useful assistant retrieves information across complex, multi-step workflows, synthesizes it into artifacts (documents, presentations, spreadsheets), and navigates program interfaces to produce a coherent final product. Such workflows demand reasoning and synthesis, decomposition of complex tasks, as well as visual and spatial understanding. To study agents on workflows like these, we introduce \textbf{\datasetname}, a benchmark of open-ended, complex, browser-based tasks that jointly evaluate these capabilities, with each task culminating in a produced artifact. To write tasks, we develop a task design rubric and a protocol for ensuring that tasks meet the requirements. Each task is paired with an \emph{evaluator}, a program that combines deterministic checks with LLM judgments to balance the richness, reliability, and automation tradeoff inherent to agent evaluation. 
We evaluate and analyze frontier computer-use agents and browser-based harnesses.
They achieve moderate scores on partial-success metrics, but the best performer fully succeeds in fewer than 3\% of our complex, long-horizon tasks. Failures on visual steps render the resulting artifacts unusable, even when agents complete more than 50\% of other evaluation steps. Our results expose limitations of current agents acting as end-to-end assistants, and call for progress on tool use, visual understanding, and long-horizon reasoning.

\end{abstract}

\section{Introduction}

Amid a rapid deployment of agents as assistants in real-world workflows, a growing body of agent benchmarks has emerged ~\citep{yehudai2026survey, kapoor2026holistic}. 
A convenient setup for evaluation is to produce tasks where success can be measured with a single, short, verifiable answer %
~\citep{yoran-etal-2024-assistantbench, wei2025browsecompsimplechallengingbenchmark}. 
However, this is a poor match for the open-ended nature of real-world tasks. 
Deep research benchmarks ~\citep{du2026deepresearch, han2026deerbenchmarkevaluatingdeep} move closer to this setting by evaluating long-form textual reports, but in the real workspace, a plain textual report is rarely the final deliverable. 
Instead, findings are arranged into artifacts (slides, spreadsheets, and documents), whose effectiveness hinges on visual, spatial, and structural choices such as the placement and color of content, the selection of accompanying images, and the organization of tabular data. 
Properly executing these choices, on top of search and synthesis, makes the full task longer-horizon, and agents that excel at producing textual outputs may fail as assistants when these last-mile steps are also required.

\begin{figure*}[t]
    \centering
    \includegraphics[width=\linewidth]{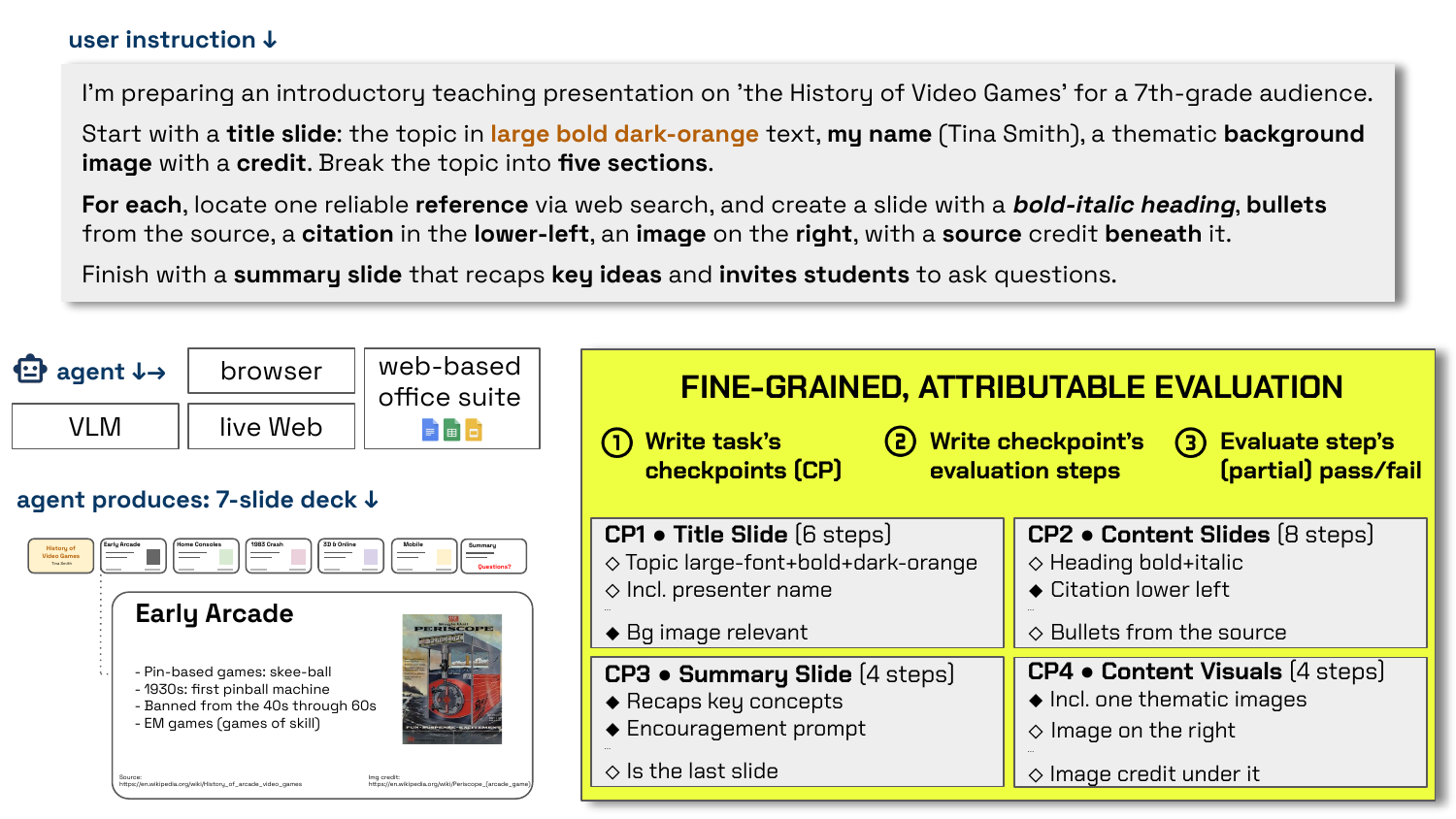}
    \caption{An overview of the \textbf{\datasetname} benchmark.}
    \label{fig:figure1}
\end{figure*}

To evaluate vision-and-language agents on these longer-horizon, open-ended tasks, we introduce \datasetname (\datasetnamefull), the first benchmark spanning live web search, productivity tool use, and visual/spatial understanding (see \sect{sec:related_work} for a comparison with existing benchmarks). Figure~\ref{fig:figure1} shows the overall setup. The agent receives a natural language instruction, uses a live web browser, a vision-and-language model (VLM), and a web-based office suite (specifically, Google Workspace) to produce a structured, coherent artifact (doc, sheet, slides) according to the user's specifications.

The more realistic, open-ended nature of the \datasetname tasks opens up a fundamental problem in agents: evaluation. In evaluating agents, we typically have a fundamental tradeoff between richness, reliability, and automation. LLM-as-a-judge evaluations ~\citep{zheng2023judging} scale, but are not fully reliable ~\citep{lu2025agentrewardbench,khalifa2026gamingjudgeunfaithfulchainofthought}, at least not without sacrificing richness by constraining the space of possible outputs. 
One approach is to leave outputs unconstrained and tackle reliability directly with fine-grained evaluation protocols ~\citep{arora2025healthbenchevaluatinglargelanguage, caldwell2025pentestjudgejudgingagentbehavior, chen2026presentbench, sharma2025researchrubrics, han2026deerbenchmarkevaluatingdeep}, but even then, reliably judging open-ended agent outputs remains difficult ~\citep{wang2026timereflecttrustllm,shen2026rethinkingrubricgenerationimproving}. 

We, therefore, use a hybrid approach to balance this tradeoff with \emph{evaluators}---programs that judge agent outputs using a combination of deterministic checks and LLMs. Consider the example in Fig.~\ref{fig:figure1}. 
The slide deck must satisfy criteria spanning layout (citation in the lower left), formatting (bold italic), source faithfulness (bullet points supported by the cited URL), and content relevance. 
We decompose each task into checkpoints, each checkpoint into evaluation steps, and check each step independently. Some steps, such as whether the image matches the theme, require a carefully prompted VLM or LLM judge. Others, like slide ordering or title color, can be checked deterministically with an API. 
The hybrid approach combines the richness of LLM judgment with the robustness we expect from deterministic checks.

\begin{figure*}[t]
    \centering
    \includegraphics[width=0.9\linewidth]{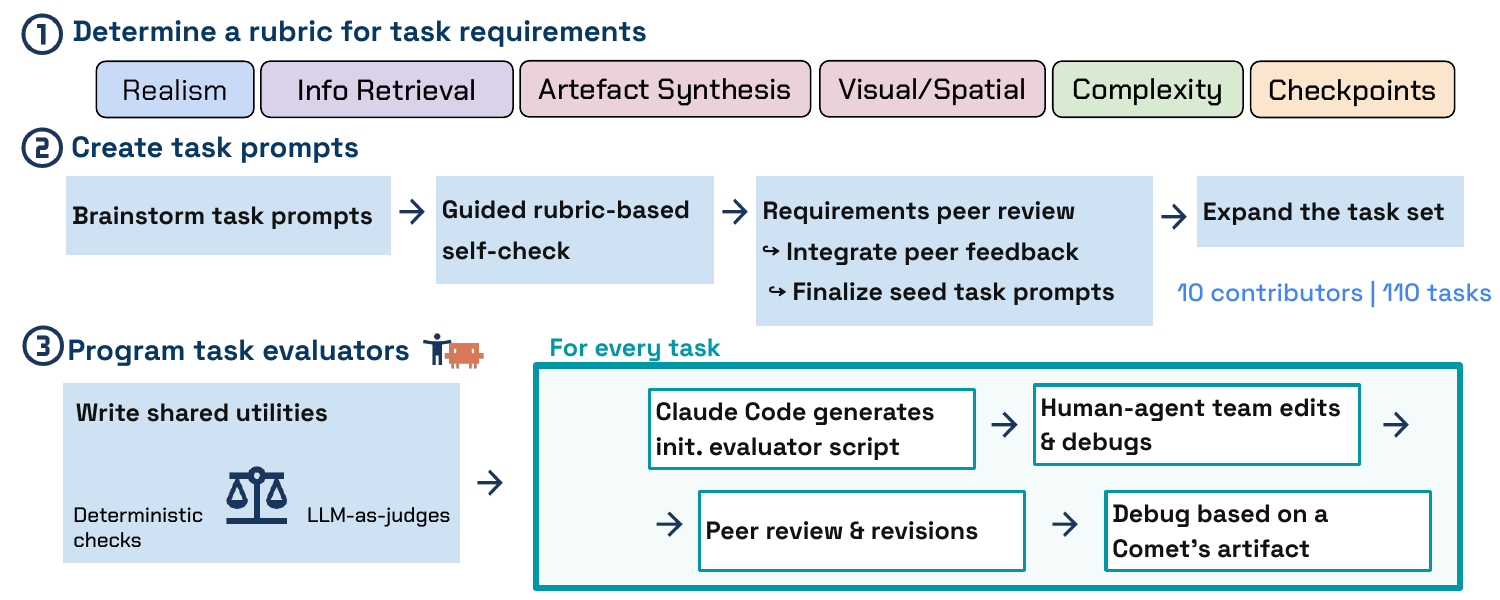}
    \caption{An overview of the construction of the \datasetname benchmark.}
    \label{fig:data-construct}
\end{figure*}

Figure~\ref{fig:data-construct} overviews how we construct \datasetname. We first define a rubric specifying the requirements for a valid task in this benchmark: realism, information retrieval, artifact synthesis, visual/spatial understanding, and complexity. We then write an initial set of tasks. %
For quality control, each task is reviewed by another annotator against the rubric, and the task is revised in response. We write additional tasks inspired by the seed set, for a total of 110 tasks. %
Each task is paired with an evaluator, which is reviewed by at least two of the authors. One author then additionally validates the scripts on real artifacts by running an agentic browser on the task instruction, scoring the resulting artifact with the current scripts, inspecting the per-step verdicts for false positives and false negatives, and the scripts are revised accordingly.

We evaluate state-of-the-art (SOTA) computer-use agents across open-source and proprietary VLMs and harnesses, and find that even the strongest one achieves only ${<}$3\% success rate on \datasetname tasks. 
Since complete success may underestimate agent capabilities, we report three partial-success metrics. Partial scores range from $\approx$35--70\% for the best-performing baseline, but we show that the resulting artifacts are often unusable. Finally, our analysis shows that proprietary harnesses outperform open-source harnesses, and that agents fail more often on steps that require source fidelity and visual/spatial understanding.
These failures call for advances in visual and spatial understanding of VLMs, long-horizon reasoning, open-source, and tool use. 
We publicly release \datasetname to the community. The project page is at
\url{https://alexgill321.github.io/KNOWS-benchmark/}, task data is at
\url{https://huggingface.co/datasets/utahnlp/knows-benchmark} and the
evaluation code at \url{https://github.com/alexgill321/KNOWS-benchmark}.

\section{Related Work}
\label{sec:related_work}
\noindent\textbf{Web agent benchmarks} such as WebShop~\citep{yao2022webshop}, WebArena~\citep{zhou2024webarena}, Mind2Web~\citep{deng2023mind2webgeneralistagentweb}, and others~\citep{koh2024visualwebarenaevaluatingmultimodalagents,lu2024weblinxrealworldwebsitenavigation, song2025bearcubs} focus mostly on relatively simple and easily verifiable tasks. These generally test an agent's ability to perform basic tasks on websites, such as navigating to a page and finding a specific piece of information. They can typically be completed in a relatively small number of actions and do not involve creating complex artifacts or synthesizing large amounts of information, as in \datasetname.

\paragraph{}As agents became more capable, agent benchmarks also involve more open-ended tasks. Many of these are categorized as \textbf{deep research benchmarks}. Datasets such as DeepResearchBench~\citep{du2026deepresearch}, Mind2Web 2~\cite{gou2025mind2web}, and DEER~\citep{han2026deerbenchmarkevaluatingdeep} require agents to retrieve information from multiple websites and combine them, much as in \datasetname. However, unlike in \datasetname, these do not have the additional step of taking the results of this research and then sorting, analyzing, and synthesizing them into structured information. While they are required to reason from the information to produce a final answer (which can often be quite long), this does not necessitate manipulating web tools nor the kinds of spatial, visual, or structured reasoning required to create the outputs required in \datasetname.

\paragraph{}Several datasets have tackled the task of \textbf{evaluating agents on productivity/office} suites. For instance, early OS benchmarks such as OSWorld~\cite{xie2024osworldbenchmarkingmultimodalagents} have required agents to do simple tasks in LibreOffice. OSWorld-MCP~\cite{jia2026osworldmcp} includes similar spreadsheet operations and formatting tasks, text editing and formatting, slide manipulation, and media insertion. Prior work has built benchmarks specific to Microsoft Office~\cite{wang2024officebench}, %
spreadsheets~\cite{ma2024spreadsheetbench} %
or documents~\cite{li2026clawsbench}, testing agents' ability to manipulate these programs, but none are complex or require notable amounts of information retrieval. OdysseyBench~\cite{wang2025odysseybench} is likely the most complex of these. However, this again mostly tests the ability to do multi-workflow office tasks rather than retrieving and synthesizing information. Our tasks, while requiring manipulation of these kinds of programs, also require information retrieval from the web, reasoning, and generating complex and coherent artifacts in these programs. 

\paragraph{}Most similar to ours is PresentBench~\citep{chen2026presentbench}, which, like some of our tasks, synthesizes slideshows. Their benchmark focuses on the specific problem of creating large, impressive presentations. However, their information retrieval is more limited (it retrieves from a curated set of academically rigorous sources), while we focus on broader information that also requires navigation of the live web. Our benchmark also focuses on more structural constraints (such as where images should be placed), whereas theirs focuses more on evaluating open-ended presentation generation.

\paragraph{}One of our main contributions to the area of computer use agents is not only the dataset and tasks themselves, but our unique way of \textbf{agentic evaluation}, balancing the difficulty of having realistic and open-ended tasks with the practicalities of evaluation. Benchmarks with simple, easy to verify outputs ~\citep[e.g., check if a particular item is in your cart; ][]{yao2022webshop} use simple automatic verification. For more ambiguous or difficult to programmatically define benchmarks, many papers use LLM-as-a-judge~\cite{zheng2023judging} or VLM-as-a-judge~\cite{chen2024mllm}, most prominently for web agents in Online-Mind2Web~\cite{xue2025an}. As tasks have gotten more difficult, evaluations have started to add intermediate checkpoints~\cite{pan2024webcanvas} or use detailed rubrics~\cite{chen2026presentbench, sharma2025researchrubrics, han2026deerbenchmarkevaluatingdeep} to evaluate tasks. \datasetname takes a hybrid approach, combining programmatic verifiers and LLM-as-a-judge, and breaks tasks into multiple discrete checkpoints and separate evaluation steps. This approach allows both more robust verification of complex tasks that can be divided into easily verifiable steps and the completion of more open-ended tasks.

\section{\datasetname}
The aim of \datasetname is to create tasks that test agents on realistic and complex artifact synthesis tasks. To this end, we follow a careful, time-intensive, iterative design process, overviewed in Figure~\ref{fig:data-construct}. We start by laying out a rubric for task creation (\S\ref{sec:desiderata}). Next, we describe our task collection process, from designing a diverse set of high-quality seed tasks to using them to write a larger task set (\S\ref{sec:taskcreation}). Then we describe how we build evaluators (\S\ref{sec:evaluators}) to score agents on these complex tasks. Finally, we present and analyze our dataset
(\S\ref{sec:datasetanalysis}).

\subsection{Task Desiderata}
\label{sec:desiderata}

We translate our core design requirements into a rubric that guides task creation and keeps new tasks aligned with our motivation. 

The first requirement, \textbf{realism}, follows recent web agent benchmarks. Tasks must be something a real user might want to do and would save time if automated. The next requirement, significant \textbf{information retrieval}, follows deep research benchmarks. Tasks must require browsing the live web with at least five (but often more!) pieces of information drawn from at least two websites. The next requirement, \textbf{artifact synthesis}, mandates that tasks produce a cohesive, well-organized artifact using a Google Workspace tool (slides, docs, or sheets). The agent solving our tasks must meaningfully synthesize retrieved information rather than just copy it directly into artifacts unedited. 
Our tasks also require some kind of \textbf{visual or spatial} component, such as identifying or editing specific elements, observing spatial relationships (left, right, up), or color-based modification (e.g., remove all red links). We again follow recent trends towards more \textbf{complexity} by requiring at least four steps of information retrieval, multiple steps to create the artifacts, and at least ten steps to incorporate and synthesize information. Finally, we require discrete \textbf{checkpoints} which break each task into measurable milestones in the creation of the artifact.

\paragraph{Google Workspace.}
We build our artifact synthesis tasks on Google Workspace (Drive, Docs, Sheets, and Slides). We choose this platform because it is used by a huge portion of internet users, supports realistic task scenarios in almost any domain, and provides a platform with API support for programmatic evaluations. 

\paragraph{Live Web.}
Despite the known challenges with live web benchmarks ~\citep{akkil2026emergencewebvoyagerconsistenttransparent, sun2025webarxivevaluatingmultimodalagents}, we nevertheless focus our information retrieval on live web. It is more realistic as there are many possible sources of information, whereas a hosted intranet requires artificially limiting the set of available sources and therefore the breadth and realism of the tasks. Deep research benchmarks, which also look to evaluate agent performance in finding information on the web \citep{du2026deepresearch, gou2025mind2web, han2026deerbenchmarkevaluatingdeep}, all use live web as well in order to get the variety and richness of sources and distractors to test realistic information retrieval. Caching or freezing a subset of pages would remove the information retrieval challenge of \datasetname by artificially limiting the set of potential sources. Internet Archive, an alternative to caching, has issues with rate limiting of web-search agents and broken content for niche sites, so we avoid its use as well.\footnote{\url{https://archive.org/}}

Using the live web does create challenges for evaluation, particularly around websites changing with time. To account for this, we aim to curate tasks that involve information extraction that is independent of a single website and time-agnostic, meaning that the information required to complete a task can be found in multiple websites and will not change based on the date it is accessed. Since we nevertheless cannot guarantee that our evaluation scripts will remain functional indefinitely, we adopt a maintenance and deprecation policy for detecting and remedying outdated tasks, detailed in Appendix~\ref{app:fut-maintenance}.

\subsection{Task Creation}
\label{sec:taskcreation}
Each task in \datasetname is composed of the following. First is the \textbf{user instruction} or \textbf{task prompt}, which is a detailed description in English of what the agent has to do. We aim for the task prompt to be clear and descriptive enough so that a detail-oriented human would be able to complete the task without further instructions. Next, each task contains a set of  \textbf{checkpoints}. Checkpoints are discrete, measurable milestones for evaluating these tasks, each containing multiple separate evaluation steps. Tasks can be scored on overall completion or partial completion, based on the number of checkpoints and evaluation steps achieved. Finally, and most importantly, each task has an \textbf{evaluator}, a script that automatically checks the artifact at each checkpoint using a combination of hard-coded scripts, utility functions such as Google Workspace APIs, and LLMs and VLMs to grade the checkpoint. See
Fig.~\ref{fig:figure1} for a task example. 

Using the rubric of task requirements, we write a Google Colab notebook (see Figures~\ref{fig:colab1}, \ref{fig:colab2}, \& \ref{fig:colab3} in Appendix~\ref{app:colab})
to guide annotators through task creation. The notebook consists of the following steps. 
It first offers brainstorming suggestions for task prompts by displaying the Google Workspace template categories and subcategories, but it leaves an option to diverge from these. 
We allow annotators any tools at their disposal, including using LLMs, to assist in the brainstorming process. 
The notebook then urges annotators to check their task prompt against each of the required desiderata, and revise the task if any are not met. 
Finally, annotators write descriptions of the checkpoints and evaluation steps within these checkpoints. %

While the workflow of self-checking each task requirement is part of the notebook, to further ensure that all tasks meet our requirements, another person reviews each task independently. They read, going back through the rubric, and verify that each is satisfied. In cases where they were not met, or there was an ambiguity or disagreement, the second annotator either edited or returned to the original annotator to either clarify or edit the task.

Once this initial set of tasks was collected, for additional coverage, we took these seed tasks and created additional tasks from them. For instance, we can take a task about preparing a presentation on the History of Video Games for a 7th-grade audience in Fig.~\ref{fig:figure1} to a presentation on CRISPR-Cas9 Gene Editing for a 10th-grade audience, keeping many of the described steps the same, but drastically changing the websites and information in the task to create very different final outputs. In total, we collected \numtotaltasks tasks.

Because of the complexity and length of the task creation pipeline (requiring in-person explanation and demonstration), these steps were done by authors 
whose work was then validated by other authors to ensure high-quality. We estimate that the creation of each task from start to finish takes around 9--18 hours of expert work to complete on average.

\subsection{Evaluators}
\label{sec:evaluators}
The final and most important step in task creation is writing evaluators. For each evaluation step in a task
there must be a verifier for the successful completion of that step implemented by an evaluator. Each task has a dedicated evaluator script, which programmatically checks completion of each evaluation step, and returns the completion scores for each checkpoint (defined below). These evaluators are implemented by examining Google Workspace tool states, analyzing agent trajectories, and extracting data from the live web.

We first note that many evaluation steps share common requirements in implementation, so we create a suite of shared evaluation utilities available to any evaluator.\footnote{Similar to evaluation programs in \citet{10.5555/3666122.3666387}.} We organize these utilities around seven categories of evaluation step. \textbf{Structural checks} verify the existence, order, and hierarchy of elements via the Google Workspace API. \textbf{Information retrieval checks} compare extracted data values against gold-standard references using exact matching, fuzzy matching, or numerical comparison with tolerances. \textbf{Formatting checks} verify visual styling properties such as font weight, size, color, and alignment by querying the API directly. \textbf{Spatial checks} validate the physical layout of elements through their position, size, alignment, and whether they overlap or extend beyond the page. \textbf{Content checks} assess source fidelity and semantic accuracy of generated text. \textbf{Web visit checks} confirm that the agent actually visited relevant websites by inspecting its trace of visited URLs. \textbf{Visual checks} verify image content and provenance---whether an inserted image depicts the expected subject, originates from the correct source, or is visually distinct from other images. 

Some checks that cannot be statically verified require careful use of LLM-as-a-judge \citep{zheng2023judging} or VLM comparison with formatting and rubrics such that they can be robustly evaluated. Additionally, for some deterministic checks, we use an LLM/VLM-as-judge as a backup check if the deterministic path fails. We describe the evaluation step categories in greater detail in Appendix~\ref{app:evaluatordetails}.

Next, we create an \textbf{evaluator} for each task. Each evaluator iterates through the task's checkpoints and verifies the success of every evaluation step within them, drawing on any of the seven evaluation check categories described above. As writing evaluators can be quite difficult and time-consuming, we use the following process. First a human creates a ``gold standard'' artifact for the task, acting as an ideal agent. 
We take the full task prompt and the codebase for the human-written evaluation utilities and feed these to Claude Code, along with an example of an evaluation script; the full prompt is given in the Appendix Figure~\ref{fig:claude-code-prompt}. This generates an initial evaluator script that, unfortunately, always contains mistakes or misses steps. The evaluator's human creator edits and debugs the evaluator script against the gold artifact. One person reviews the evaluator and requests fixes to the evaluation script until it is fully correct on the gold artifact and approved.
However, even then, bugs can persist.  
We therefore inspect how the evaluators score artifacts made by Comet \citep{yang2025adoptionusageaiagents}, document the remaining issues, and finalize the evaluators with these final issues resolved. We provide more information on evaluator contributors in Appendix \ref{app:eval-cont-det}.

\paragraph{Evaluator Validation.} To measure the reliability of the finalized evaluators, an expert annotator, who did not implement the evaluators, judged whether artifacts sampled across baseline agents satisfy individual evaluation steps, for a total of 100 judgments. We find that evaluators agree with expert judgment with a Cohen's $\kappa$ of 0.64 and 82\% pairwise accuracy, in line with related open-ended agent benchmarks \citep{xue2025an, du2026deepresearch, han2026deerbenchmarkevaluatingdeep}. We also validate evaluator--expert agreement on an independent set of 100 evaluator decisions consisting entirely of LLM/VLM-based steps. We find agreement with a Cohen's $\kappa$ of 0.69 and 89\% pairwise accuracy on these judgments.

\input{tables/datasetstats}

\subsection{Benchmark Overview}
\label{sec:datasetanalysis}

The \datasetname dataset contains \numtotaltasks tasks across three artifact types: 25 documents, 40 slides, and 45 spreadsheets. Task examples are given in the Appendix Table~\ref{tab:task-prompts}. For a randomly chosen task, we present our decomposition of the task into checkpoints and of each checkpoint into evaluation steps in the Appendix Figure~\ref{fig:task-decomposition}. We provide \datasetname's overall statistics in Table~\ref{tab:dataset_stats}. One distinguishing feature is the length of these tasks. We see in Table~\ref{tab:dataset_stats} that the queries themselves for tasks are quite long (200 words on average) and the estimated time to complete the tasks (Appendix Table~\ref{tab:human-completion-times}) is over 2 hours for humans. Tasks also span a broad range of domains. We use a taxonomy from Google Workspace use-case categories (\S\ref{sec:taskcreation}), with our tasks covering 11 domains from the original taxonomy and an additional 9 use cases extending it. We list the full taxonomy and per-domain task counts in Appendix~\ref{app:domain-taxonomy}.

We also report the distribution of evaluation step types in Table~\ref{tab:eval_templates}, which we calculated by prompting Claude Opus 4.6 to categorize each step's natural language description based on our definitions.\footnote{The full prompt is given in the Appendix Figure~\ref{listing:cat_prompt}.} %

\input{tables/mainBenchmark}

\section{Experimental Results}

In this section, we introduce evaluation metrics (\sect{sec:eval-metrics}), baseline agents (\sect{sec:baselines}), and finally, show how they perform on \datasetname (\sect{sec:results}).

\subsection{Evaluation Metrics}
\label{sec:eval-metrics}
Let a task instance $i$ consist of $C_i$ checkpoints, where checkpoint $j$ contains $S_{ij}$ evaluation steps, of which $k_{ij}$ are correct. We define four metrics:

\noindent\textbf{Success Rate (SR)} is the standard metric for \textit{complete success}, where a task instance scores $1$ only if every evaluation step across all checkpoints is correct. We define success rate for a task $i$ as
\begin{equation}
    \text{SR}_i = \mathbb{I}\!\left[\sum_j k_{ij} = \sum_j S_{ij}\right].
\end{equation}

Given the difficulty of our tasks, SR alone is too coarse. We introduce three \textit{partial success} metrics, ordered from \textit{strictest} to \textit{loosest}.

\noindent\textbf{Average Successful Checkpoints (ASC)} assigns $1$ to a checkpoint if \emph{all} of its evaluation steps are correct, and averages across the checkpoints:
\begin{equation}
    \text{ASC}_i = \frac{1}{C_i}\sum_{j=1}^{C_i} \mathbb{I}\!\left[k_{ij} = S_{ij}\right].
\end{equation}
Since each checkpoint represents a measurable milestone in artifact creation, ASC shows how many milestones an agent can fully complete.

\noindent\textbf{Average Checkpoint Fraction (ACF)} evaluates steps within each checkpoint, then averages the fraction across checkpoints, formulated as
\begin{equation}
    \text{ACF}_i = \frac{1}{C_i}\sum_{j=1}^{C_i} \frac{k_{ij}}{S_{ij}}.
\end{equation}
ACF is similar in spirit to ASC but less strict. Rather than requiring full checkpoint completion, it awards partial credit based on how much of each checkpoint was completed.

\noindent\textbf{Step Fraction (SF)} is simply the fraction of correct evaluation steps for each task:
\begin{equation}
\text{SF}_i = {\displaystyle\sum_{j=1}^{C_i} k_{ij}}/{\displaystyle\sum_{j=1}^{C_i} S_{ij}}.
\end{equation}

SF is similar in strictness to ACF, but by aggregating evaluation steps rather than checkpoints, it naturally weights complex checkpoints (with more evaluation steps) more heavily.

\subsection{Baselines}
\label{sec:baselines}
We benchmark SOTA proprietary models: Claude Opus 4.7~\cite{claude-opus-4.7-system-card}, GPT-5.5~\cite{gpt-5.5-system-card}, and an open-source model: DeepSeek V4 Pro~\cite{deepseek2026v4}. We benchmark models with the open-source BrowserGym-AgentLab harness \cite{chezelles2025the} across two modes: text-only using the accessibility tree (AXT) and multimodal using both screenshots (SS) and AXT.\footnote{At the time of writing, visual inputs were not available for 
the official DeepSeek V4 Pro API.} We also evaluate closed-source harnesses (often called AI browsers): ChatGPT Atlas~\cite{openai2025atlas} using a GPT agent,\footnote{The specific agent is auto-selected by the harness based on the task.} and Perplexity's Comet \cite{yang2025adoptionusageaiagents}, using a Claude Opus 4.7 agent. More details are in Appendix~\ref{sec:appendix-exp-details}. We do not report a controlled human baseline, as a single independent annotator per-task would cost roughly \$5000 in labor alone (Appendix~\ref{app:human-completion-times}). Every task was, however, completed by an author during construction, and resulting artifacts pass all evaluator criteria (\S\ref{sec:taskcreation}), so agent failures we observe are not artifacts of unsolvable tasks.

\subsection{Results and Analysis}
\label{sec:results}

Table~\ref{tab:mainBenchmark} shows the \textit{full} and \textit{partial} success across three splits: \texttt{Docs}, \texttt{Sheets}, and \texttt{Slides}, and the \texttt{Overall} performance.

\paragraph{Long-horizon tasks remain challenging.} 
SR is near zero across most methods and splits, showing both the difficulty of our benchmark and the inability of SOTA agents on challenging, long-horizon tasks. {The exception is \texttt{Docs}, where the best baseline, Comet, achieves a non-zero, albeit low, SR (12.0\%).} This result is unsurprising, as doc tasks resemble the conversational, text-only outputs best suited for the underlying LLMs.

\begin{figure}[t]
    \centering
    \begin{adjustbox}{cfbox=black 1pt}
    \includegraphics[width=\linewidth]{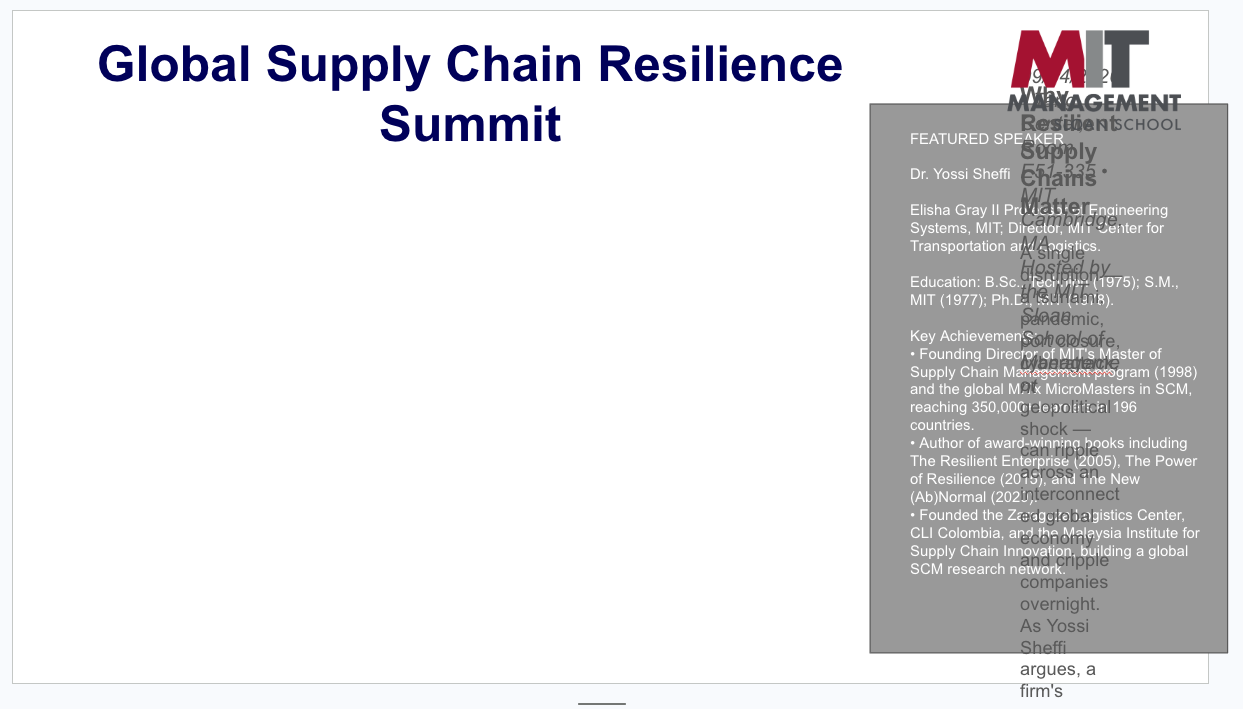}
    \end{adjustbox}
    \caption{Output produced by the best-performing agent, Comet, on a \texttt{Slides} task. It received moderate partial scores, ACF (0.54) / SF (0.59), but the output remains practically unusable (indicated by a low ASC of 0.14).}
    \label{fig:cometError}
\end{figure}

Partial success is also similarly low across most methods. Among the general agents, GPT-5.5 performs relatively well compared to other models. While Comet achieves the best performance, qualitative results show many of these outputs are often unusable, such as the one in Figure~\ref{fig:cometError}.

\input{figs/crossTabs-joint}

\paragraph{Visual modality is important.} \datasetname contains a large fraction of tasks that require visual and spatial reasoning, which are often unsolvable from the AXT alone. {Adding visual inputs as screenshots (SS) with the AXT can significantly improve performance, as much as ($+10.3$ ASC, $+25.5$ ACF, $+25.5$ SF) for Opus 4.7 over its AXT-only variant.}

\paragraph{Harness matters as much as the model.} Perhaps the starkest finding is the gap between two systems sharing the same
backbone, e.g., {Opus 4.7 with the Comet harness outperforms the BrowserGym harness by (+2.7 SR, +18.7 ASC, +27.5 ACF, and +25.6 SF) overall.} Proprietary harnesses show clear advantages in live web browsing, e.g., on complex webpages and with pop-up advertisements.

\paragraph{Step-wise performance analysis.} 
Because of our hybrid evaluation framework, failures are directly traceable in \datasetname. Each failed evaluation step terminates under a specific condition in its evaluator script, which we automatically map to an error category.
Fig.~\ref{fig:comet_breakdown} breaks down the performance of the best agent, Comet: Fig.~\ref{fig:category_steps_breakdown} by the mechanism that decides each step's verdict, and Fig.~\ref{fig:category_steps_breakdown_eval} by the type of checks the step evaluates (\sect{sec:datasetanalysis}). 
The mechanism types are \textsc{Exact Match} comparisons (e.g., string equality), \textsc{Tolerance Match} (e.g., fuzzy string matching), \textsc{Model Judgment} (e.g., VLM-determined content relevance), \textsc{Geometric} (e.g., bounding-box overlap), \textsc{Tree Query} (e.g., slide structure), \textsc{Unmet Dependency} (e.g., a missing element prevents a dependent check), \textsc{Execution Error} (the check could not run, e.g., missing artifact content or an API exception), \textsc{Unchecked Pass} (benefit-of-the-doubt branches). See Table~\ref{tab:failure-category-defs} (Appendix) for detailed descriptions and examples.\footnote{\textsc{trace match} has been excluded from our analysis because Comet does not return browsing traces, and the corresponding evaluation steps were performed manually.}

{Comet is strongest on deterministic \textsc{exact match} (83.2\%), weaker on \textsc{model judgment} (68.9\%), \textsc{geometric} (65.2\%), and \textsc{tolerance match} (65.1\%) steps, and weakest on \textsc{tree query} steps (60.8\%), 84 of whose 124 failures come from a single seed task in which Comet produced a deck without the requested slide structure. The remaining \textsc{unmet dependency} and \textsc{execution error} failures arise before the intended check runs, typically because content it depends on is missing from the artifact.

\paragraph{Qualitative error analysis.} While the best-performing agents can achieve decent partial success scores, several of the resulting outputs often remain practically unusable. As seen in Figure~\ref{fig:cometError}, the agent achieves 0.54 and 0.59 in ACF and SF, respectively, while the resulting artifact is a visual mess. {This failure pattern is consistent with the weaker \textsc{geometric} and \textsc{tree query} scores in Fig.~\ref{fig:category_steps_breakdown}, which check the layout and structure of the artifact.} We observe similar errors in structuring documents (Figure~\ref{fig:cometDocsError}), applying formulae in spreadsheets, structuring tables, and creating proper visualizations (Figure~\ref{fig:cometSheetsError}).

\paragraph{Cross-platform performance.}
In real use cases, assistant agents are not only restricted to creating artifacts in one platform environment. We perform a small cross-platform check to test whether our results are due to agent or harness optimization on Google Workspace. Specifically, we have Comet, our top performing model--harness combination, complete a subset of 5 of our tasks using Microsoft 365 web apps. We then grade these attempts manually using the same checkpoints and evaluation steps. We find that Comet performs better on 2 tasks and worse on another 2 tasks. Our results indicate that performance is not driven by optimization on the Google Workspace environment. Full results are in Appendix~\ref{app:cross-platform}.

\section{Conclusions}
Our results show that current agents can often find the right information, but frequently fail to synthesize, organize and display that information into coherent, usable artifacts. \datasetname tests this gap directly, evaluating web agents on complex tasks that require all of these capabilities.

Our experiments reveal a clear capability gap. {The best performing model, Comet, passes most exact-value checks, but is weaker on checks of the fidelity, layout, and structure of the artifact (Figure~\ref{fig:category_steps_breakdown}).} No model successfully completes more than 3\% of the total \datasetname task set (Table~\ref{tab:mainBenchmark}). Even when partial success metrics suggest moderate progress, the resulting artifacts are often practically unusable (Figure \ref{fig:cometError}). Retrieval is not solved either, with Comet passing two thirds of information-retrieval checks, but the widest gaps lie in the visual understanding, tool knowledge, and reasoning required to produce coherent final products.

Our results also highlight that the agent harness matters as much, if not more, than the underlying model. Claude Opus 4.7, using two different harnesses, shows a large performance gap, pointing to infrastructure, action spaces, and tool interfaces as key areas driving and potentially hindering the performance of current web agents. We release \datasetname and all evaluation code to support future work on closing these gaps. \footnote{Project page: \url{https://alexgill321.github.io/KNOWS-benchmark/}}

\section*{Limitations}

\paragraph{Domain coverage.} While \datasetname selectively collects 110 realistic tasks, it may not cover the full breadth of domains encountered in real-world web use. One design decision in \datasetname is its focus on depth over scale. While some benchmarks, such as WebArena~\cite{zhou2024webarena} and VisualWebArena~\cite{koh2024visualwebarenaevaluatingmultimodalagents}, prioritized broader task coverage, \datasetname instead emphasized task complexity and length. Other recent benchmarks, such as Mind2Web 2~\cite{gou2025mind2web} and WorkArena~\cite{workarena2024}, share a similar philosophy. Each of our 110 tasks required approximately 9--18 person-hours to construct, including evaluator creation and debugging, reflecting the care taken to ensure each task is meaningful, well-validated, and representative of real-world complexity. This rigor also has practical implications at evaluation time, as a single baseline run takes at least 10 hours to complete (depending on the benchmarking hardware and parallelization). 

\paragraph{Live web evaluation.} The challenges of live web benchmarks have been well known in the web-use domain~\citep{akkil2026emergencewebvoyagerconsistenttransparent, sun2025webarxivevaluatingmultimodalagents}. Most significantly, this makes the benchmark depend on the current state of the web, which can change. Web changes can take the form of simple UI changes, the updating and outdating of information, and, occasionally, the loss of websites entirely. This dependence means our benchmark's validity is partially contingent on the current state of the web at evaluation time, a limitation present across all live web benchmarks.

We nevertheless believe that the live web aspect was crucial to properly evaluate information retrieval. In real information retrieval tasks, you must find a particular piece of information on an Internet-scale corpus of potential knowledge and extract the necessary information, where there may be many sources for that piece of information. Limiting to a small subset of locally hosted websites would not realistically test information retrieval. Virtually all DeepResearch-type benchmarks also make this choice~\citep{du2026deepresearch,gou2025mind2web,han2026deerbenchmarkevaluatingdeep}. To mitigate the challenges associated with live web, we design our benchmark by carefully curating tasks for which the required information is available on multiple websites and that generally do not change over time, e.g., we avoid tasks where the agent needs to find the front-page story. Our evaluation scripts pull from the live web, and we acknowledge they may break. To address this, we commit to periodically testing all scripts after release and to promptly respond to reports of outdated tasks via the project's GitHub repository (Appendix~\ref{app:fut-maintenance}).

\paragraph{Artifact scope.} Our benchmark is limited to three artifact types: Google Sheets, Slides, and Docs, leaving the broad landscape of real-world web tools untested. The resulting coverage is nevertheless broad: three artifact types spanning tabular data, visual presentations, and long-form documents support a wide range of task structures, and this diversity is comparable to, or even greater than, that of prior datasets (see \S\ref{sec:related_work}). Our artifact choice also gives models explicitly trained on Google Workspace a structural advantage over those trained on other platforms. To partially mitigate the training confound, we deliberately exclude Google's own models from our evaluation.

We additionally perform a small cross-platform check with our top performing model to determine the extent that training with Google Workspace affects our findings (\S\ref{sec:results}). A full cross-platform check is not within the scope of this benchmark due to many of the deterministic evaluator checks relying on Google API utilities to function. That said, the design of our shared evaluation utilities could be ported to extend to other platforms in future work building on \datasetname. While our evaluators rely on Google Workspace APIs, web agents completing the tasks only interact with workspace artifacts through the rendered browser UI via generic actions (Table~\ref{tab:action_space}).

\section*{Ethical Considerations}

One set of ethical considerations for \datasetname fall in the category of general risks for web agents. This benchmark looks to benchmark and push the envelope on the capabilities of agents to perform meaningful, long-term useful work. Improved agents raise a number of safety and privacy concerns, such as unsupervised agents leaking a user's personal information, spamming or performing harmful actions on websites, purchasing items without authorization, and more. This benchmark specifically gives access to a Google account, which might be connected to a user's professional or personal information.\footnote{We recommend that people evaluating on our dataset make a new account to avoid this issue.} There are real concerns about the speed of progress in the field and how much access users are giving agents before the work on security and trustworthiness has fully caught up. We evaluate tasks that do not ask the agent to send emails and messages, post on blogs, use credit cards, buy products, or generally interact with others on the web, so we may avoid many of these issues. However, continued development of these agents will inevitably bring these issues up in the future as agents become more capable.

A more specific concern relating to the kinds of tasks in this dataset is that it allows work to be automated, which might introduce issues. If, for instance, educators use agents to automate some parts of education (examples of which can be found in this dataset), this could lead to the particular biases and viewpoints inherent in these models to be reflected more broadly. It may also cause issues related to mistakes, hallucinations, lack of nuance or other pedagogical losses due to agents lacking the ability to tailor their content. Shifting tasks to agents blurs the accountability and transparency in these processes. This concern is one instance of a larger problem yet to be fully addressed: how LLM technology has shaped the education of millions around the world before society has had the time to investigate the effects of the technology or shape new norms. 

Personally Identifiable Information is another ethical consideration for this dataset. As several of the tasks involve collecting information from the web, sometimes this includes PII of individually identifiable people. All of the PII in this dataset comes directly from the live web. We do not introduce or add any new PII that was not already directly available on the web before \datasetname was created. In addition, to ensure compliance with rules and norms around PII, we contacted the owners of all PII and received explicit permission to use the information for the purpose of creating the dataset. The information involved is not of a sensitive nature, but it is the kind of information that people might put on their personal website (as indeed they do in one of our tasks), such as emails and affiliations.

\section*{Author Contributions}

 Alex Gill led the project, managing the creation of templates, instances and evaluators. Md Farhan Ishmam led the benchmarking aspects of the project. Kenneth Marino, Ana Marasovi\'{c}, and Alex Gill designed the benchmark. Detailed contributions were as follows:

\begin{itemize}
    \item \textbf{Conception:} Alex Gill, Kenneth Marino, Ana Marasovi\'{c}
    \item \textbf{Benchmark Design:} Alex Gill, Md Farhan Ishmam, Fateme Hashemi Chaleshtori, Nathan Stringham, Kenneth Marino, Ana Marasovi\'{c}
    \item \textbf{Task Creation:} Alex Gill, Md Farhan Ishmam, Fateme Hashemi Chaleshtori, Nathan Stringham, Xuyen Nguyen, Neha Bhat, Parker DeYoung, Kenneth Marino, Ana Marasovi\'{c}
    \item \textbf{Evaluator Development:} Alex Gill, Xuyen Nguyen, Neha Bhat, Parker DeYoung
    \item \textbf{Benchmarking:} Md Farhan Ishmam
    \item \textbf{Paper Writing:} Alex Gill, Md Farhan Ishmam, Kenneth Marino, Ana Marasovi\'{c}
    \item \textbf{Advising:} Kenneth Marino and Ana Marasovi\'{c} advised the project, providing guidance and substantial paper editing.
\end{itemize}

\section*{Acknowledgments}
We thank Google for providing funding for the construction of this benchmark. We thank Anthropic and OpenAI for their donation of API credits for benchmarking models. We thank Ivan Andhika, Matthew Lee, Purbid Bambroo, Ryhor Pryslopski, Spike Cheng and Qui Ngo for their contributions to the codebase, Shrusti Ghela and Ana Alvarez Lopez for providing valuable annotations, and everyone who consented to our use of their information in this benchmark. We also thank everyone involved in the review process for providing valuable feedback on \datasetname.

\bibliography{references}

\appendix

\section{\datasetname Collection}
\subsection{Task Creation Colab}
\label{app:colab}

Following our rubric of task requirements, we wrote a Google Colab notebook to
  guide annotators through task creation. For legibility, we report the
  notebook as text in Figures~\ref{fig:colab1}, \ref{fig:colab2},
  and \ref{fig:colab3}.

\subsection{Task Domain Taxonomy}
\label{app:domain-taxonomy}
We report the full taxonomy and per-domain task coverage counts in Table~\ref{tab:domain-coverage}. This table shows the Google Workspace use-case category that tasks fall under, and whether or not that use case comes from the original taxonomy or an additional use case. In total, there are 20 unique domains, each consisting of at least 5 tasks.

\subsection{Human Completion Times}
\label{app:human-completion-times}

To establish a human baseline for task difficulty, we measured the time required for human annotators to complete each task by analyzing edit activity logs from the Google Drive Activity API. For each task template, we examined the edit history of the gold document used for debugging when creating the original evaluators. Edit events are recorded at approximately three-minute intervals by the API. We segmented editing activity into sessions by treating gaps longer than one hour as session boundaries, then summed the duration of all sessions attributed to the primary annotator. We show the completion time to create a gold artifact for one instance of each seed task (Table~\ref{tab:human-completion-times}).

As discussed in Section~\ref{sec:baselines}, we do not report a controlled human baseline in our experiments. From Table~\ref{tab:human-completion-times} we estimate each task would take an average of 3 hours per task. At \$15/hr pay, 110 tasks would cost \$4950, not accounting for any platform fees, and this is with only a single annotator for each task.

\subsection{Evaluator Details}
\label{app:evaluatordetails}

Each task instance includes a programmatic evaluator that decomposes the task requirements into fine-grained evaluation steps. Many evaluation steps share common implementation requirements, so we developed a suite of shared evaluation utilities accessible to all evaluators. Below we describe the implementation techniques used for each category of evaluation step defined in Section~\ref{sec:evaluators}.

\subsubsection{Structural Checks}
Many tasks require organizing a Google Workspace document in a specific manner: creating particular columns in a spreadsheet, ordering sections in a document, or including required elements such as charts or bullet lists. Structural checks verify the existence, order, and hierarchy of these elements---not their styling or position, but whether the right pieces are present and correctly arranged. For example, the investment tracker task checks that columns exist for stock name, ticker symbol, and price, while the recipe task verifies that an ``Ingredients'' column appears first. To implement these checks, we leverage the Google Workspace API, which provides structural information with ordered sequential data for all elements (text, images, etc.) in a Workspace object. Using this, we locate elements in an object's structure and compare their position relative to other elements.

\subsubsection{Information Retrieval Checks} Since every task requires live web information retrieval, a substantial portion of evaluation steps verify that the agent retrieved correct data. These steps compare values in the completed document against gold-standard reference data, checking that numbers, names, dates, and other facts match expected values within appropriate tolerances. For example, the investment tracker task verifies that stock prices match current market data, and the food composition task checks that nutritional values match USDA database entries. Implementation relies on deterministic comparisons: exact string matching, fuzzy text matching, and numerical comparison with configurable error tolerances.

\subsubsection{Formatting Checks} These steps verify visual styling properties of document elements: font weight, size, and color, cell background colors, text alignment, merged cells, and other presentation attributes. For example, the event poster task checks that the event name is bold, at least 20pt, and black, while the food composition task verifies that column headers are bolded and group headers are centered and italicized. These checks query styling properties directly through the Google Workspace API and are fully deterministic.

\subsubsection{Content Checks} While information retrieval checks verify exact data values, content checks assess whether information is faithfully tied to the correct source, whether summaries accurately represent source material, or whether claims are properly substantiated. For example, the event poster task checks that a speaker biography is ``clearly tied to'' the speaker's field of expertise and that cited sources correspond to specific claims. Because these judgments require semantic understanding, content checks are the category most reliant on LLM-as-a-Judge~\citep{zheng2023judging}. We format each requirement as a binary yes/no question and use an LLM to assess whether the content meets the criterion, following the robust performance of LLMs on binary classification~\citep{Gilardi_2023}.

\subsubsection{Spatial Checks} While structural checks verify element ordering within the document structure, spatial checks validate an element's physical position in the rendered document as seen by a human, such as where something appears on the page, how large it is, or its spatial relationship to other elements. For example, ``a logo must be in the top right of the slide'' or ``no text or images are overlapping each other.'' For Google Slides and Sheets, we extract element positions directly from the Google Workspace API, which provides each element's translation offset and size in English Metric Units (EMUs). We convert these to bounding boxes and use geometric operations such as overlap ratios, containment checks, and center-point comparisons to evaluate spatial relationships. For example, to verify that a text box appears at the top of a slide, we check whether its vertical offset falls within the top 20\% of the slide height. To verify that elements do not overlap, we compute pairwise bounding box intersections. For Google Docs, where the API does not expose positional coordinates, we render PDF exports as images at 300 DPI and use scale-invariant template matching~\citep{lowe1999object} to locate specific elements within the rendered page.

\subsubsection{Web Visit Checks} As discussed in Section~\ref{sec:desiderata}, each task is designed to require live web information retrieval. However, LLMs may memorize correct information during pre-training, potentially enabling agents to complete tasks without actually browsing the web. To verify that agents perform genuine web retrieval, we require a list of all URLs visited during task execution to be passed to our evaluators. We do not require that specific URLs be visited, since our tasks leave the choice of website open to the agent. Instead, we pull HTML data from each visited URL and determine whether the information in the completed Workspace document could have been extracted from the set of visited pages. This verification uses either exact match or LLM-as-a-Judge, determined on a case-by-case basis depending on the complexity of the information. For some tasks, the agent is required to cite source links in the completed artifact. In those cases, we verify that the sources appear in their browsing history.

\subsubsection{Visual Checks} These steps verify that a specific image is present in the document or that an image matches expected content. For example, the Wikipedia photos task checks that ``the wiki image of Tom Hanks is present on the slide,'' and the paper sorting task verifies that generated figures match gold-standard chart images. Implementation varies by task: some steps use deterministic image comparison (pixel-level matching or perceptual hashing), while others require a VLM to identify whether an image depicts the expected content.

\subsection{Evaluator Contributor Details}
\label{app:eval-cont-det}
Students from the authors' university helped develop the evaluator scripts. We recruited them through a mix of word-of-mouth and advertising the project in class. Task prompts themselves were written by the authors, while contributors worked exclusively on implementing, debugging and evaluating the per-task evaluator scripts against gold artifacts. Contributors participated through one or more of the following arrangements: research-assistant role paid at \$20/hr, course credit for research, or paid hourly work at \$15/hr.

\subsection{Benchmark Maintenance and Deprecation Policy}
\label{app:fut-maintenance}

Live web tasks may be affected by changes or updates to websites over time. We intentionally limit the susceptibility of \datasetname to website changes by aiming to curate tasks that are time-agnostic and independent of any single website. Nevertheless, we cannot guarantee that every task and evaluation script will remain functional indefinitely, so we adopt the following maintenance and deprecation policy, modeled after the policy of Mind2Web 2 \citep{gou2025mind2web}. 

We have curated a list of 23 URLs and 20 tasks with website-extracted gold data that our evaluators rely on, and we commit to manually checking these at periodic intervals after release. We prioritize reliability over ease of running these checks, and therefore commit to manual verification. In parallel, we will validate automatic freshness checks against the manual ones, and will switch to automated checking once it proves consistently accurate.

If substantial website changes or unavailability significantly alter a task's difficulty or solvability, we will update the affected task or replace it with a new one of similar complexity and scope. Any changes to tasks or evaluators will be logged as a new version of the benchmark on GitHub, and we will archive each prior version so that results reported against it remain interpretable and reproducible. We include instructions on the project's GitHub repository \url{https://github.com/alexgill321/KNOWS-benchmark} for community reporting of any outdated tasks or evaluator issues.

\begin{table*}[ht]
\centering
\small
\begin{tabular}{@{}llccc@{}}
\toprule
Category & Use Case & Modalities & Tasks & Source \\
\midrule
Letters   & Business Letter               & Docs         & 5  & Taxonomy \\
Personal  & Recipes                       & Docs, Sheets & 10 & Taxonomy \\
Education & Lesson Plan                   & Docs, Slides & 10 & Taxonomy \\
Education & Class Notes / Reference List  & Docs         & 5  & Taxonomy \\
Education & Research Report               & Docs         & 5  & Taxonomy \\
Education & Book Report                   & Slides       & 5  & Taxonomy \\
Education & Research Library              & Sheets       & 5  & Extension \\
Personal  & Investment Tracker            & Sheets       & 5  & Taxonomy \\
Personal  & Travel Planner                & Sheets       & 5  & Taxonomy \\
Personal  & Wedding Planner               & Sheets       & 5  & Taxonomy \\
Personal  & Fitness Analytics             & Sheets       & 5  & Extension \\
Personal  & Outdoor Activity Planner      & Sheets       & 5  & Extension \\
Personal  & Housing Search                & Sheets       & 5  & Extension \\
Personal  & Media Recommendation          & Sheets       & 5  & Extension \\
Personal  & Lookbook                      & Slides       & 5  & Taxonomy \\
Personal  & Party Invite / Event Poster   & Slides       & 5  & Taxonomy \\
Personal  & Product Comparison            & Slides       & 5  & Extension \\
Personal  & Purchase Decision             & Slides       & 5  & Extension \\
Work      & Photo Directory               & Slides       & 5  & Extension \\
(Cross)   & Slide Editing                 & Slides       & 5  & Extension \\
\midrule
20 unique domains & & & 110 & 11 T / 9 E \\
\bottomrule
\end{tabular}
\caption{Domain coverage of the \datasetname benchmark.
Each row is a distinct use-case domain identified by (category, use case). Domains are drawn from a Google Workspace use-case taxonomy defined in the task creation colab notebook (Appendix~\ref{app:colab}). 11 fall within the pre-defined taxonomy (``Taxonomy'') and 9 are extensions we introduced to capture live web-heavy use cases the original design under-specified (``Extension''). Every domain is represented by at least 5 individual tasks via our template$\times$instance structure.}
\label{tab:domain-coverage}
\end{table*}

\begin{table}[ht]
\centering
\small
\resizebox{0.5\textwidth}{!}{%
\begin{tabular}{ll}
\toprule
\textbf{Action} & \textbf{Description} \\
\midrule
\texttt{click}                & Single left-click on an element \\
\texttt{dblclick}             & Double-click on an element \\
\texttt{drag\_and\_drop}      & Drag an element to a target location \\
\texttt{fill}                 & Clear and fill a text input \\
\texttt{go\_back}             & Navigate to the previous page \\
\texttt{go\_forward}          & Navigate to the next page \\
\texttt{goto}                 & Navigate to a given URL \\
\texttt{hover}                & Move cursor over an element \\
\texttt{keyboard\_insert\_text} & Insert text at the cursor position \\
\texttt{keyboard\_press}      & Press a key or key combination \\
\texttt{keyboard\_type}       & Type text character by character \\
\texttt{mouse\_click}         & Click at absolute screen coordinates \\
\texttt{mouse\_dblclick}      & Double-click at screen coordinates \\
\texttt{mouse\_drag\_and\_drop} & Drag between two screen coordinates \\
\texttt{mouse\_move}          & Move mouse to screen coordinates \\
\texttt{mouse\_upload\_file}  & Upload a file via mouse interaction \\
\texttt{new\_tab}             & Open a new browser tab \\
\texttt{noop}                 & Perform no operation \\
\texttt{report\_infeasible}   & Flag the task as infeasible \\
\texttt{scroll}               & Scroll the page or an element \\
\texttt{select\_option}       & Select an option from a dropdown \\
\texttt{send\_msg\_to\_user}  & Send a message to the user \\
\texttt{tab\_close}           & Close the current tab \\
\texttt{tab\_focus}           & Switch focus to a specific tab \\
\texttt{upload\_file}         & Upload a file to an input element \\
\bottomrule
\end{tabular}
}
\caption{Action space of \datasetname.}
\label{tab:action_space}
\end{table}

\section{Experimental Details}
\label{sec:appendix-exp-details}
Our tasks require the creation of new Google Workspace artifacts, and the evaluation scripts directly use the Workspace API for evaluation. To run this benchmark, we first authenticate with a Google account and, for each task instance, the framework creates blank output artifacts to be filled in by agents. Some tasks also bundle supplementary input data---such as reference images, recipe PDFs, or CSV files---in a local \texttt{data/} directory accessible to the agent. Once baselines are run, we then use these created artifacts for our evaluator scripts (see \S\ref{sec:evaluators}) to score agent performance.

\subsection{Live Web Harness}
For evaluating our LLM baselines, we use the BrowserGym ecosystem~\cite{chezelles2025the}, which provides an easy common interface for web browsing tasks. This not only provides the infrastructure, but also a standard interface for observation spaces including accessibility trees and action spaces, which allows for both low-level actions (e.g., $\texttt{click}$) as well as more general, high-level interactions (e.g., $\texttt{goto}$). We report the action space in Table~\ref{tab:action_space}. Some baselines, such as Comet~\citep{yang2025adoptionusageaiagents} are integrated into browsers and are thus pre-packaged with their own harness, and we use these instead.

\subsection{Google Authentication}
\label{subsec:googleAuth}
To avoid bot detection, Google authentication state is captured fresh before each task and stored per worker \texttt{PID}, giving each parallel Ray worker its own isolated Playwright storage state. Shared snapshots and persistent Chromium profiles are deliberately avoided to maintain worker isolation. Should a session expire mid-task, inline re-authentication instructions serve as a fallback.

\subsection{Evaluation Cost}
Here we report the monetary cost of running our baselines on the \datasetname tasks. Token- and step-level efficiency of baselines cannot be reported uniformly across all agents because Comet and Atlas are consumer products that don't expose token consumption or action traces, and are priced as flat-rate subscriptions. For Comet and Atlas, wall-clock time is not directly comparable as those agents will occasionally stop to request user assistance. While we do not provide any assistance in our benchmarking, we do need to prompt them to continue the task with no assistance in these instances. The one metric that is comparable across every system is monetary cost, which we report in Table~\ref{tab:cost}.

\subsection{Quantitative Error Analysis}
\label{app:failure-categories}
Every evaluator in \datasetname{} has verdicts produced by explicit checks. We categorize all scoring call sites across evaluator instances so that each step records which type of mechanism decided a pass/fail verdict, along with the verdict itself. We fix mechanism categorizations under the conditional branches an evaluation step terminates at, so assignment is always deterministic.

\paragraph{Categories.}
Table~\ref{tab:failure-category-defs} defines the scoring mechanism taxonomy. Six categorizations can occur on failure or pass verdicts, while {\textsc{unchecked pass}} occurs only on passing steps, and \textsc{unmet dependency} and \textsc{execution error} occur only on failing steps.

\paragraph{Assignment rules.}

Many evaluation steps have multiple fallback checks (e.g., exact pixel match $\rightarrow$ perceptual hash $\rightarrow$  VLM comparison). In these cases, the step records the last mechanism that ran, since it makes the final call on the step's success verdict. When an LLM extracts the value under test, but a deterministic or tolerance-based comparison renders the verdict, the step records the comparator's category. When a single step scores $N$ sub-items whose verdicts may come from different mechanisms, the step records the majority category among the failing items if any item failed, and the majority category among all items otherwise.

\paragraph{Step-wise analysis details.}
We grade the artifacts produced by the best performing baseline, Comet, on all 110 tasks. We exclude the 189 web-visit steps from the mechanism analysis, since every one of them requires the browsing trace, leaving 2{,}508 steps. Comet passes 1{,}706 steps (68.0\%) and fails 802. {Figure~\ref{fig:category_steps_breakdown} summarizes the distribution.} Table~\ref{tab:step-mechanism-harness} breaks pass rates and failure shares down by artifact type, and Table~\ref{tab:failure-attribution} attributes the failures. Below, seed tasks are referred to by artifact type and number (e.g., Docs-11), matching the identifiers of our released tasks. Three observations stand out: (a) the same mechanism behaves very differently across artifact types: \textsc{tree query} steps are near-perfect in \texttt{Sheets} (97\%, column and row existence and ordering) but the weakest mechanism in \texttt{Slides} (42\%, required slides present and in order) and \texttt{Docs} (45\%, section layout of recipe pages and reference lists), (b) \textsc{geometric} checks barely exist outside \texttt{Slides}, where they are a genuine weakness (56\%), (c) \texttt{Docs} is the only artifact type in which \textsc{exact match} failures dominate (30.4\% of its failures).

\begin{table}[h]
\centering\small
\begin{tabular}{llr}
\toprule
System & Configs & Total \\
\midrule
\multicolumn{3}{l}{\emph{Usage-based (API)}} \\
Claude Opus 4.7 & AXT, AXT+SS & \$2{,}200 \\
GPT-5.5 & AXT, AXT+SS & \$1{,}700 \\
DeepSeek V4 Pro & AXT & \$80 \\
\midrule
\multicolumn{3}{l}{\emph{Flat-rate (subscription)}} \\
Comet & --- & \$200 \\
Atlas & --- & \$100 \\
\bottomrule
\end{tabular}
\caption{Cost of one full evaluation pass over all 110 tasks. Totals are not
comparable across rows: the API systems differ in the number of
configurations run, and subscription tiers are billed monthly and are
independent of usage volume.}
\label{tab:cost}
\end{table}

\paragraph{Failure attribution.}
Table~\ref{tab:failure-attribution} attributes Comet's failures over all 2{,}697 steps. Of the 939 failed steps, 667 (71.0\%) reflect a check that ran on the artifact and rejected it. A further 114 (12.1\%) are web-visit steps that cannot be scored because no browsing trace was available for that task. The remaining 158 (16.8\%) are \textsc{execution error} or \textsc{unmet dependency} steps where the intended check never ran, most often because content it requires is missing from the artifact.

\input{tables/stepMechanismAppendix}

\subsection{Cross-Platform Full Results}
\label{app:cross-platform}
Here we report full results for 5 task artifacts created in Microsoft Office by Comet, as discussed in \S\ref{sec:results}. Table~\ref{tab:cross-platform} shows the results comparing the scores for Comet artifacts created on the same tasks using both Google Workspace and Microsoft Office as the platforms.

\begin{table*}[t]
\centering
\small
\begin{tabular}{lp{0.62\linewidth}}
\toprule
Category & Deciding mechanism \\
\midrule
\textsc{Exact Match} & Exact or normalized comparison: string equality, regular expressions, exact numeric equality, pixel-exact image match, API-reported formatting flags. \\
\textsc{Tolerance Match} & Tolerance-based, non-geometric comparison: fuzzy string matching, perceptual image hashing, numeric comparison within a relative or absolute tolerance, color-distance thresholds. \\
\textsc{Model Judgment} & An LLM or VLM verdict decided the step (content relevance, semantic equivalence, image--description agreement). \\
\textsc{Geometric} & A geometric test over rendered coordinates decided the
step: bounding-box containment or overlap, page-region position, area coverage, OCR-localized text position. \\
\textsc{Tree Query} & An artifact structural test decided the outcome of a step: element
counts, ordering, position within the document tree, slide or column structure. \\
\textsc{Trace Match} & A string or pattern match over the agent's browsing history decided the step. These check that the agent visited the websites a task requires.  \\
\textsc{Unmet Dependency} & The step was deemed a failure because a
prerequisite step failed (e.g., a location check cannot be performed because the
corresponding text is missing from the artifact). Failures in this category are cascades,
not independent errors. \\
\textsc{Execution error} & The check could not run because its input could not be obtained: a required part of the artifact could not be located or parsed (e.g., no slides in the deck, header row not detected), an evaluator input such as the browsing trace or reference data was unavailable, or an external lookup or API call failed. \\
\midrule
\textsc{Unchecked Pass} & (Success only.) The step passed without a substantive check, e.g., nothing to verify, an unreachable source, or missing reference data. \\
\bottomrule
\end{tabular}
\caption{Failure-category taxonomy. Each evaluation step records the category
of the mechanism that decided its outcome.}
\label{tab:failure-category-defs}
\end{table*}

\begin{table*}[t]
\centering
\begin{tabular}{lccr}
\toprule
\textbf{Task} & \textbf{MS 365} & \textbf{Google Workspace} & \textbf{$\Delta$} \\
\midrule
docs\_1\_formal\_letter                        & 8/10    & 8/10    & $0$   \\
docs\_31\_education\_lesson\_plan              & 99/145  & 132/145 & $-33$ \\
slides\_29\_buy\_car\_pres                     & 57/68   & 62/68   & $-5$  \\
slides\_42\_personal\_none\_product\_comparison & 43/57   & 26/57   & $+17$ \\
slides\_51\_event\_announcement\_poster        & 175/185 & 100/185 & $+75$ \\
\midrule
\textbf{Combined} & \textbf{382/465 (82\%)} & \textbf{328/465 (71\%)} & $+54$ \\
\bottomrule
\end{tabular}
\caption{Cross-platform comparison of agent performance on five \datasetname tasks
implemented in both Microsoft 365 and Google Workspace. Scores are evaluation steps scores out of total points available for each task. $\Delta$ is the difference in points earned (MS $-$ GWS).}
\label{tab:cross-platform}
\end{table*}

\begin{figure*}[ht!]
    \centering
    \begin{adjustbox}{cfbox=black 1pt}
        \includegraphics[width=\linewidth]{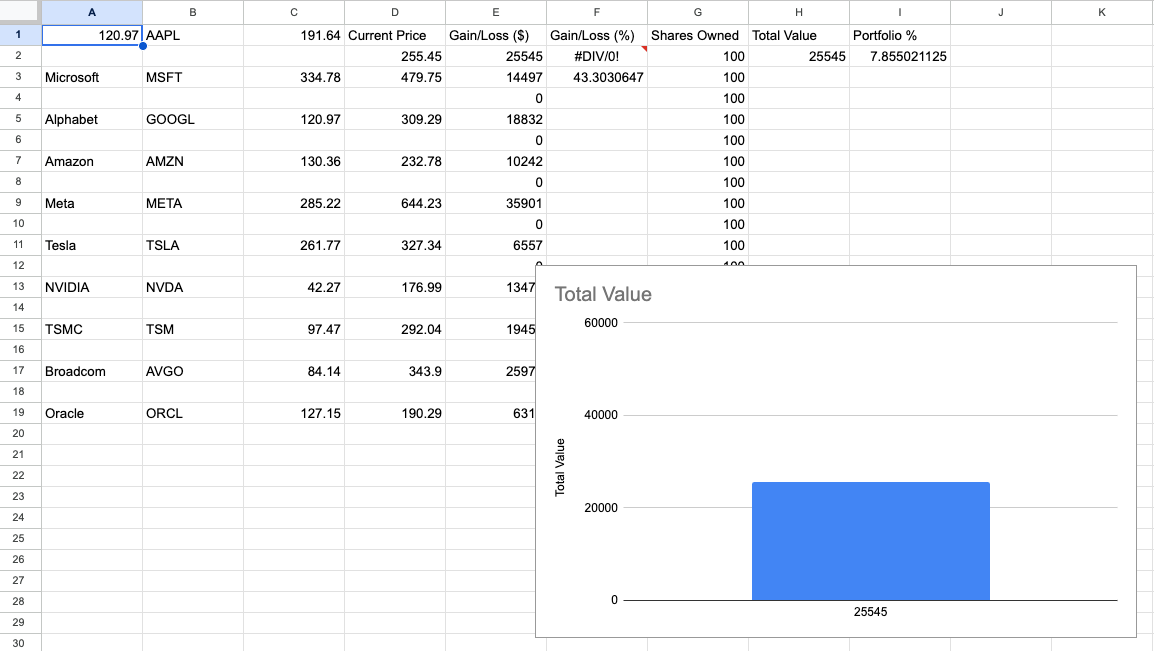}
    \end{adjustbox}
    \caption{Output produced by the best performing agent, Comet, on a \texttt{Sheets} task. The agent makes several errors, including formula errors, inconsistent table structure, misplaced formulas, and an inability to generate visualizations correctly.}
    \label{fig:cometSheetsError}
\end{figure*}

\begin{figure}[!ht]
    \centering
    \begin{adjustbox}{cfbox=black 1pt}
        \includegraphics[width=\linewidth]{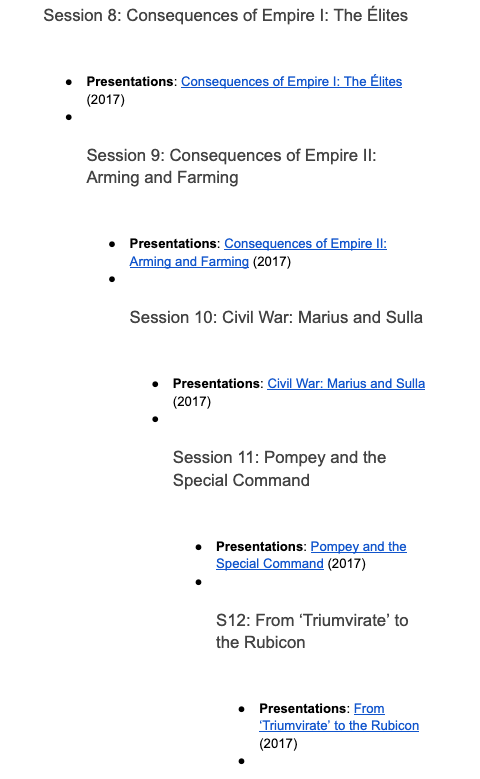}
    \end{adjustbox}
    \caption{Output produced by the best performing agent, Comet, on a \texttt{Docs} task. The agent generates deeply nested bullet points, making the document difficult to navigate and unpresentable.}
    \label{fig:cometDocsError}
\end{figure}

\input{tables/human_gold_instance_time}

\input{prompts/colab_listings}

\input{tables/task_decomposition_example}
\input{tables/task_prompts_table}

\input{prompts/claude_code_prompt}
\input{prompts/step_categorization}
\end{document}

%% file: tables/datasetstats.tex
\begin{table}[t]
  \centering
  \setlength{\tabcolsep}{8pt}
  \begin{tabular}{lc}
    \toprule
    \textbf{Statistic} & \textbf{Value} \\
    \midrule
 
    Total Tasks  & 110 \\[4pt]
 
    \multicolumn{2}{l}{\textit{Prompt length}} \\[1pt]
    \quad Words (mean / median)      & 204.4 / 165.0 \\
    \quad Sentences (mean / median)  & 13.7 / 12.0  \\[2pt]
 
    \multicolumn{2}{l}{\textit{Checkpoints per task}} \\[1pt]
    \quad Mean / Median  & 5.9 / 5.0 \\
    \quad Min -- Max     & 4 -- 9   \\[2pt]
 
    \multicolumn{2}{l}{\textit{Evaluation steps per task}} \\[1pt]
    \quad Mean / Median  & 24.6 / 23.0 \\
    \quad Min -- Max     & 9 -- 70     \\
    \bottomrule
  \end{tabular}
    \caption{%
    Core statistics of the \datasetname dataset.
  }
    \label{tab:dataset_stats}

\end{table}
 
\bigskip
 
  \begin{table}[t]
    \centering
    \setlength{\tabcolsep}{4pt}
    \renewcommand{\arraystretch}{1.15}
    \begin{tabular}{@{}lrrr@{}}
      \toprule
      \textbf{Step Type}
        & \textbf{\# Steps}
        & \textbf{\% Steps}
        & \textbf{\% Inst.} \\
      \midrule
      Structural        &   835 &  30.7 &  90.9 \\
      Info. Retrieval   &   748 &  27.5 & 100.0 \\
      Formatting        &   332 &  12.2 &  78.2 \\
      Content           &   229 &   8.4 &  54.5 \\
      Spatial           &   199 &   7.3 &  59.1 \\
      Web Visit         &   189 &   7.0 &  68.2 \\
      Visual            &   184 &   6.8 &  63.6 \\
      \midrule
      \textbf{Total}    & 2{,}716 & 100.0 &  \\
      \bottomrule
    \end{tabular}
    \caption{%
    Distribution of types of checks \datasetname steps evaluate. \% Steps is the share of all 2{,}716 evaluation steps; \% Inst.\ is the share of 110 tasks containing at least
    one step of that type.
  }
  \label{tab:eval_templates}
\end{table}

%% file: tables/mainBenchmark.tex
\begin{table*}[t]
\setlength{\tabcolsep}{2.5pt}
\centering
\normalsize
\resizebox{\textwidth}{!}{%
\begin{tabular}{l crrr | crrr | crrr | crrr}
\toprule
\multirow{3}{*}{\textbf{Method}}& \multicolumn{4}{c}{\textbf{Docs}} & \multicolumn{4}{c}{\textbf{Sheets}} & \multicolumn{4}{c}{\textbf{Slides}} & \multicolumn{4}{c}{\textbf{Overall}} \\
\cmidrule(lr){2-5} \cmidrule(lr){6-9} \cmidrule(lr){10-13} \cmidrule(lr){14-17}
 & SR & ASC & ACF & \multicolumn{1}{c}{SF} & \multicolumn{1}{c}{SR} & ASC & ACF & \multicolumn{1}{c}{SF} & SR & ASC & ACF & \multicolumn{1}{c}{SF} & SR & ASC & ACF & \multicolumn{1}{c}{SF} \\
\midrule
\multicolumn{17}{c}{Textual (Accessibility Tree)} \\
\midrule
\multicolumn{1}{l|}{\includegraphics[height=2.75ex, valign=c]{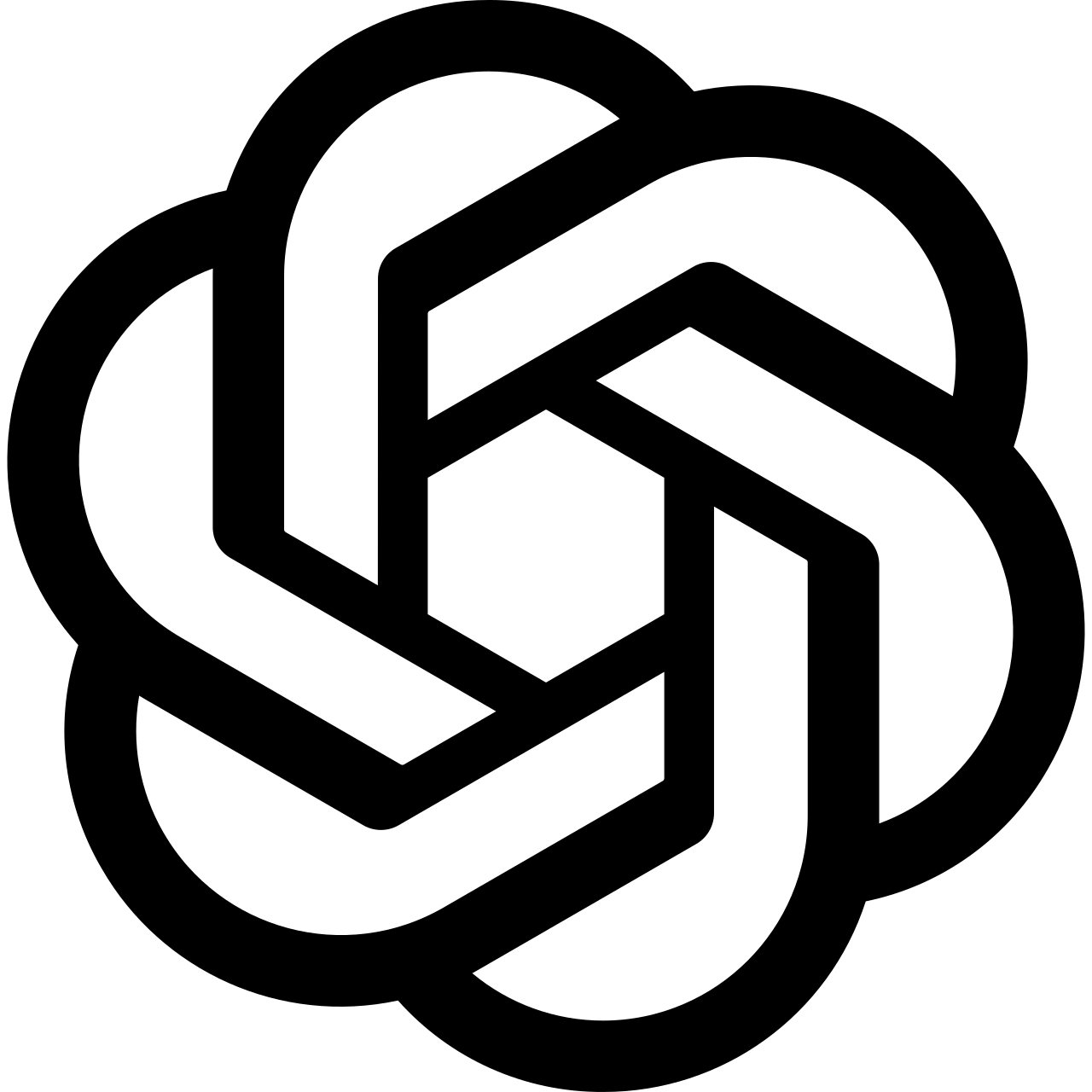}\hspace{0.2em}GPT-5.5}          & 0 & 18.0 & 38.5 & 36.6 & 0 & 26.8 & 46.8 & 44.4 & 0 & 15.1 & 38.6 & 31.7 & 0 & 20.6 & 41.9 & 38.0 \\
\multicolumn{1}{l|}{\includegraphics[height=2.75ex, valign=c]{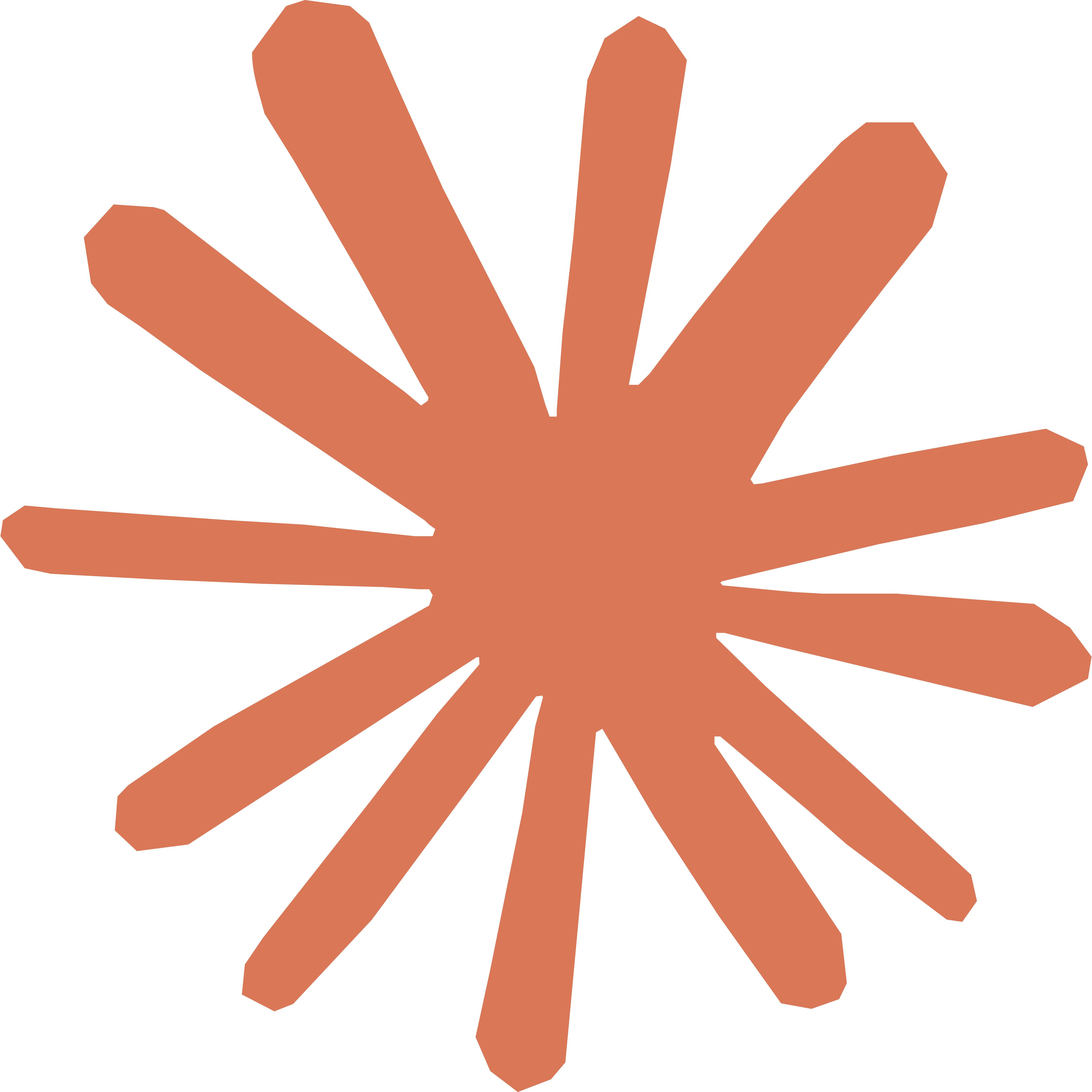}\hspace{0.2em}Claude Opus 4.7}          & 0 & \multicolumn{1}{r}{7.7} & 19.9 & 21.9 & 0 & \multicolumn{1}{r}{3.0} & 12.3 & 10.0 & 0 & \multicolumn{1}{r}{9.3} & 20.3 & 12.5 & 0 & \multicolumn{1}{r}{6.4} & 17.0 & 13.6 \\
\multicolumn{1}{l|}{\includegraphics[height=2.75ex, valign=c]{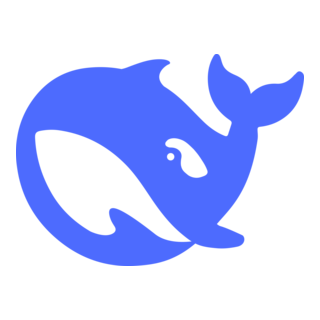}\hspace{0.2em}DeepSeek V4 Pro}          & 0 & 17.0 & 39.3 & 41.1 & 0 & \multicolumn{1}{r}{8.7} & 23.7 & 18.6 & 0 & 14.2 & 34.3 & 26.1 & 0 & 12.6 & 31.1 & 26.4 \\
\midrule
\multicolumn{17}{c}{Multimodal (Accessibility Tree + Screenshot)} \\
\midrule
\multicolumn{1}{l|}{\includegraphics[height=2.75ex, valign=c]{logo/gpt.png}\hspace{0.2em}GPT-5.5}   & 0 & 21.7 & 44.9 & 46.2 & 0 & 25.0 & 50.1 & 48.2 & 0 & 18.1 & 51.7 & 45.3 & 0 & 21.8 & 49.5 & 46.7 \\
\multicolumn{1}{l|}{\includegraphics[height=2.75ex, valign=c]{logo/claude.png}\hspace{0.2em}Claude Opus 4.7}   & 0 & 21.3 & 47.3 & 49.0 & 0 & 11.4 & 28.8 & 24.8 & 0 & 19.7 & 54.8 & 48.9 & 0 & 16.7 & 42.5 & 39.1 \\
\midrule
\multicolumn{17}{c}{AI Browsers} \\
\midrule
\multicolumn{1}{l|}{\includegraphics[height=2.75ex, valign=c]{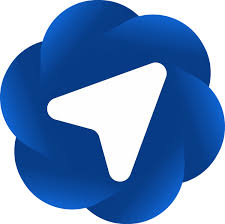}\hspace{0.2em}ChatGPT Atlas}           & 0 & 22.3 & 43.6 & 40.6 & 0 & 25.9 & 51.5 & 48.6 & 0 & 16.3 & 39.7 & 31.6 & 0 & 21.6 & 45.4 & 40.6 \\
\multicolumn{1}{l|}{\includegraphics[height=2.75ex, valign=c]{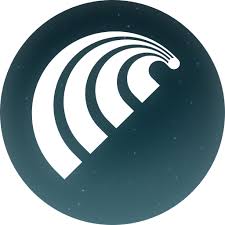}\hspace{0.2em}Perplexity Comet}           & \multicolumn{1}{r}{\cellcolor{cyan!07}\textbf{12.0}} & \cellcolor{cyan!07}\textbf{41.0} & \cellcolor{cyan!07}\textbf{68.7} & \cellcolor{cyan!07}\textbf{66.3} & 0 & \cellcolor{cyan!07}\textbf{35.8} & \cellcolor{cyan!07}\textbf{73.0} & \cellcolor{cyan!07}\textbf{68.3} & 0 & \cellcolor{cyan!07}\textbf{31.3} & \cellcolor{cyan!07}\textbf{67.5} & \cellcolor{cyan!07}\textbf{59.7} & \multicolumn{1}{r}{\cellcolor{cyan!07}\textbf{2.7}} & \cellcolor{cyan!07}\textbf{35.4} & \cellcolor{cyan!07}\textbf{70.0} & \cellcolor{cyan!07}\textbf{64.7} \\
\bottomrule
\end{tabular}%
}
\caption{Results on the \datasetname dataset. SR, ASC, ACF, and SF are metrics described in \sect{sec:eval-metrics}, with values reported as percentages (\%). Baseline agents that are \emph{not} AI Browsers use the BrowserGym-AgentLab harness.}
\label{tab:mainBenchmark}
\end{table*}

%% file: figs/crossTabs-joint.tex
\begin{figure}[t]
    \centering
    \begin{subfigure}[t]{\linewidth}
            \vspace{0pt}          %
        \centering
    \begin{tikzpicture}
    \begin{axis} [
        ybar stacked,
        width=\columnwidth,
        height=4.5cm,
        enlarge y limits={upper, value=0.22},
        nodes near coords bar offset=1,  %
        enlarge x limits=0.07,
        ylabel={\# Steps},
        ylabel shift=-5pt,
        ymin=0,
        ytick={0, 250, 500, 750, 1000, 1250},
        grid=both,
        major grid style={dashed, gray!55},
        minor grid style={white},
        minor tick num=1,
        symbolic x coords={em, tm, mj, ge, tq, ud, ee, up},
        legend image code/.code={
            \draw[#1, draw=black] (0cm,-0.12cm) rectangle (0.15cm,0.12cm);
        },
        xtick=data,
        xticklabels={Exact match, Tolerance match, Model judgment, Geometric, Tree query, Unmet dependency, Execution error, Unchecked pass},
        xticklabel style={rotate=45, anchor=east, font=\footnotesize, xshift=2pt, yshift=-2pt},
        bar width=11pt,
        legend style={
            at={(0.85,0.65)},
            anchor=south,
            legend columns=1,
            font=\small
        },
        tick label style={font=\small},
    ]
    \addplot[fill=magenta!30, draw=black,
        postaction={pattern=north east lines, pattern color=black!90}]
        coordinates {
        (em, 750)
        (tm, 226)
        (mj, 410)
        (ge, 118)
        (tq, 192)
        (ud,   0)
        (ee,   0)
        (up,  10)
    };
    \addplot[fill=teal!30, draw=black]
        coordinates {
        (em, 151)
        (tm, 121)
        (mj, 185)
        (ge,  63)
        (tq, 124)
        (ud,  78)
        (ee,  80)
        (up,   0)
    };

    \node[font=\footnotesize, anchor=south, yshift=1pt] at (axis cs:em,901) {83.2};
    \node[font=\footnotesize, anchor=south, yshift=1pt] at (axis cs:tm,347) {65.1};
    \node[font=\footnotesize, anchor=south, yshift=1pt] at (axis cs:mj,595) {68.9};
    \node[font=\footnotesize, anchor=south, yshift=1pt] at (axis cs:ge,181) {65.2};
    \node[font=\footnotesize, anchor=south, yshift=1pt] at (axis cs:tq,316) {60.8};
    \node[font=\footnotesize, anchor=south, yshift=1pt] at (axis cs:ud,78) {0.0};
    \node[font=\footnotesize, anchor=south, yshift=1pt] at (axis cs:ee,80) {0.0};
    \node[font=\footnotesize, anchor=south, yshift=1pt] at (axis cs:up,10) {100.0};
    \legend{Passed, Failed}
    \end{axis}
    \end{tikzpicture}
    \caption{W.r.t.\ the type of mechanism that decided each step's verdict (Appendix~\ref{app:failure-categories}), excluding the 189 web-visit steps, which require Comet's browsing trace.}
    \label{fig:category_steps_breakdown}
    \end{subfigure}
    
    \vspace{1em}
    
    \begin{subfigure}[t]{\linewidth}
        \centering
        \vspace{0pt}
    \begin{tikzpicture}
    \begin{axis} [
        ybar stacked,
        width=\columnwidth,
        height=4.5cm,
        enlarge y limits={upper, value=0.2},
        enlarge x limits=0.10,
        ylabel={\# Steps},
        ylabel shift=-5pt,
        ymin=0,
        ytick={0, 250, 500, 750, 1000, 1250},
        grid=both,
        major grid style={dashed, gray!55},
        minor grid style={white},
        minor tick num=1,
        symbolic x coords={sc, ir, cc, fc, spc, vc, wvc},
        legend image code/.code={
            \draw[#1, draw=black] (0cm,-0.12cm) rectangle (0.15cm,0.12cm);
        },
        xtick=data,
        xticklabels={Structure, IR, Content, Format, Spatial, Visual, Web Visit},
        xticklabel style={rotate=30, anchor=east, yshift=-6pt},
        bar width=14pt,
        legend style={
            at={(0.85,0.65)},
            anchor=south,
            legend columns=1,
            font=\small
        },
        tick label style={font=\small},
    ]

    \addplot[fill=magenta!30, draw=black,
        postaction={pattern=north east lines, pattern color=black!90}]
        coordinates {
        (sc, 684)
        (ir, 497)
        (cc, 138)
        (fc, 192)
        (spc, 126)
        (vc,  69)
        (wvc,  52)
    };

    \addplot[fill=teal!30, draw=black]
        coordinates {
        (sc, 144)
        (ir, 249)
        (cc,  91)
        (fc, 130)
        (spc,  73)
        (vc, 115)
        (wvc, 137)
    };

    \node[font=\footnotesize, anchor=south, yshift=1pt] at (axis cs:ir,746) {66.6};
    \node[font=\footnotesize, anchor=south, yshift=1pt] at (axis cs:sc,828) {82.6};
    \node[font=\footnotesize, anchor=south, yshift=1pt] at (axis cs:cc,229) {60.3};
    \node[font=\footnotesize, anchor=south, yshift=1pt] at (axis cs:fc,322) {59.6};
    \node[font=\footnotesize, anchor=south, yshift=1pt] at (axis cs:spc,199) {63.3};
    \node[font=\footnotesize, anchor=south, yshift=1pt] at (axis cs:vc,184) {37.5};
    \node[font=\footnotesize, anchor=south, yshift=1pt] at (axis cs:wvc,189) {27.5};

    \legend{Passed, Failed}
    \end{axis}
    \end{tikzpicture}
    \caption{W.r.t.\ the types of checks steps evaluate; see \sect{sec:datasetanalysis}.}
\label{fig:category_steps_breakdown_eval}
    \end{subfigure}
    \caption{Breakdown of Comet's verdicts across all 110 task instances: (a) on 2{,}508 steps, excluding web-visit steps; (b) on all 2{,}697 steps. Bar height is the number of evaluation steps, split into passed (hatched) and failed steps. The label above each bar gives the pass rate (\%).}
    \label{fig:comet_breakdown}
\end{figure}

%% file: tables/stepMechanismAppendix.tex
\begin{table*}[t]
\centering
\footnotesize
\setlength{\tabcolsep}{4pt}
\resizebox{\textwidth}{!}{%
\begin{tabular}{l rrrr | rrrr}
\toprule
\multirow{2}{*}{\textbf{Step type}} & \multicolumn{4}{c|}{\textbf{Passed / total steps}} & \multicolumn{4}{c}{\textbf{Share of failed steps (\%)}} \\
\cmidrule(lr){2-5} \cmidrule(lr){6-9}
 & Docs & Sheets & Slides & All & Docs & Sheets & Slides & All \\
\midrule
\textsc{exact match}      & 74/129 (57\%)  & 500/566 (88\%)     & 176/206 (85\%) & 750/901 (83\%)   & 30.4 & 23.7 &  8.7 & 18.8 \\
\textsc{tolerance match}  & 43/60 (72\%)   & 130/207 (63\%)     & 53/80 (66\%)   & 226/347 (65\%)   &  9.4 & 27.7 &  7.9 & 15.1 \\
\textsc{model judgment}  & 65/97 (67\%)   & 188/280 (67\%)     & 157/218 (72\%) & 410/595 (69\%)   & 17.7 & 33.1 & 17.8 & 23.1 \\
\textsc{geometric}        & 20/20 (100\%)  & 28/36 (78\%)       & 70/125 (56\%)  & 118/181 (65\%)   &  0.0 &  2.9 & 16.0 &  7.9 \\
\textsc{tree query}       & 23/51 (45\%)   & 101/104 (97\%)     & 68/161 (42\%)  & 192/316 (61\%)   & 15.5 &  1.1 & 27.1 & 15.5 \\
\textsc{unmet dependency} & 0/7            & 0/26               & 0/45           & 0/78             &  3.9 &  9.4 & 13.1 &  9.7 \\
\textsc{execution error}  & 0/42           & 0/6                & 0/32           & 0/80             & 23.2 &  2.2 &  9.3 & 10.0 \\
\textsc{unchecked pass}   & 6/6            & 3/3                & 1/1            & 10/10            &  0.0 &  0.0 &  0.0 &  0.0 \\
\midrule
\textbf{All}              & 231/412 (56\%) & 950/1{,}228 (77\%) & 525/868 (60\%) & 1{,}706/2{,}508 (68\%) & \multicolumn{4}{c}{Failed: 181 / 278 / 343 / 802} \\
\bottomrule
\end{tabular}%
}
\caption{Comet's evaluation steps by scoring mechanism and artifact type, excluding the 189 web-visit steps. Left: passed / total steps (pass rate). Right: share of each artifact type's failed steps decided by each mechanism. Last row gives number of failed steps.}
\label{tab:step-mechanism-harness}
\end{table*}

\begin{table*}[t]
\centering
\footnotesize
\begin{tabular}{lrrrrr}
\toprule
\textbf{Failure class} & \textbf{Steps} & \textbf{\%} & \textbf{Docs} & \textbf{Sheets} & \textbf{Slides} \\
\midrule
Check ran and rejected the artifact            &   667 &  71.0 & 138 & 256 & 273 \\
Requires the browsing trace                    &   114 &  12.1 &  10 & 104 &   0 \\
Execution error / dependency skip              &   158 &  16.8 &  49 &  32 &  77 \\
\midrule
\textbf{All failed steps}                      &   939 & 100.0 & 197 & 392 & 350 \\
\bottomrule
\end{tabular}%
\caption{Attribution of Comet's 939 failed evaluation steps. Only the first class is a verdict on the artifact. The second is unscorable without the browsing trace, and the third never reached their intended check.}
\label{tab:failure-attribution}
\end{table*}

%% file: tables/human_gold_instance_time.tex
\begin{table}[t]
\centering
\small
\begin{tabular}{@{}lr@{}}
\toprule
\textbf{Task} & \textbf{Time} \\
\midrule
docs\_5\_influential\_papers & 24m \\
slides\_20\_Illustrated\_Book\_Report & 44m \\
docs\_31\_education\_lesson\_plan & 55m \\
sheets\_6\_investmenttracker & 1.4h \\
docs\_1\_formal\_letter & 1.6h \\
slides\_39\_Personal\_Lookbook\_PaintColors & 1.6h \\
sheets\_55\_Movie\_Recommendation & 1.9h \\
sheets\_2\_personal\_recipe\_foodcomposition & 1.9h \\
docs\_11\_personal\_recipe\_ocr & 1.9h \\
sheets\_25\_skitourplan & 2.1h \\
slides\_17\_removeimagesaddplaceholders & 2.2h \\
sheets\_38\_apartment\_finder & 2.4h \\
sheets\_10\_paper\_sorting & 3.1h \\
sheets\_7\_running\_analysis & 3.1h \\
slides\_29\_buy\_car\_pres & 3.2h \\
slides\_51\_event\_announcement\_poster & 3.4h \\
slides\_30\_Work\_Wikipedia\_Photos & 3.5h \\
slides\_26\_basic\_educational\_slide\_deck & 4.1h \\
sheets\_45\_wedding\_color\_pallette & 4.4h \\
sheets\_28\_personal\_travel\_planner & 7.3h \\
docs\_37\_reference\_list & 14.5h \\
slides\_42\_personal\_none\_product\_comparison & 18.8h \\
\midrule
Median & 2.3h \\
Mean & 3.8h \\
\bottomrule
\end{tabular}
\caption{Human completion times for each task template (instance~1), measured from Google Drive Activity API edit logs. Times are sorted in ascending order.}
\label{tab:human-completion-times}
\end{table}

%% file: prompts/colab_listings.tex
\onecolumn
\captionof{figure}{The brainstorming phase of the task-creation notebook.
The annotator first picks a Google Workspace tool and a optionally (top-level,
low-level) category, then writes a brief overview and a first-draft task
prompt. They are explicitly told they may diverge from the suggested
categories.}
\begin{tcblisting}{
    breakable,
    enhanced,
    colback=gray!5,
    colframe=black!70,
    boxrule=0.6pt,
    arc=2pt,
    left=8pt, right=8pt, top=6pt, bottom=6pt,
    title=\textbf{Step 1 of the notebook: Brainstorm a task prompt},
    fonttitle=\bfseries,
    coltitle=white,
    colbacktitle=black!70,
    listing only,
    listing options={
        basicstyle=\ttfamily\footnotesize,
        breaklines=true,
        breakatwhitespace=true,
        columns=fullflexible,
        keepspaces=true,
        showstringspaces=false,
    },
}
=== Choose Type / Categories ===

> What type of template is it?
  Options: Docs (d)  Sheets (sh)  Slides (sl)
< sh
Your template type is Sheets

> Now choose the top level template type.
  Note: you are not restricted to these (select None instead) and some
  categories may be hard to write templates for.
  Options: Personal (pe)  Work (w)  Project Management (pr)
           Education (e)  None (n)
< pe

> Now choose low level category
  Options: To-do List (to)  Annual Budget (annual b)  Monthly Budget (m)
           Google Finance Investment Tracker (g)  Annual Calendar (annual c)
           Schedule (s)  Travel Planner (tr)  Wedding Planner (w)
           Team Roster (te)  Pros and Cons (p)  None (n)
< s
You have chosen: Personal | Schedule

=== Brainstorm ===

Now that you have your categories, brainstorm a bit. If your idea doesn't
quite match the categories you chose above, that's fine -- go back and
rerun those to match the idea. Think of things you might want a digital
assistant to do in your own life, or put yourself in someone else's shoes.

> Please give a brief overview of the task you are asking the agent to do.
< Make a playlist sheet for an event for my favorite artist.

=== Draft Template ===

Example Task:
  Structure a formal letter written in my name using Google Docs. Use my
  website for my info: {link}. Add the logo of my {"university",
  "department", "company", "employer"} to the upper left if available on
  the Web. Take a photo of my signature from the drive {location} and add
  it to the end of the letter.

> Write the most straightforward prompt for what you want the template to
  do. For now, don't worry about the requirements.
< I need to make a playlist for my {EVENT}. I want all of the songs to be
  from {ARTIST/BAND/COMPOSER}. First, in one tab, list at least 20 songs
  ... [annotator's draft prompt]
\end{tcblisting}
\label{fig:colab1}

\captionof{figure}{The desiderata-check phase. The notebook walks the
annotator through a True/False rubric grouped into five categories
(utility and realism, document synthesis, information retrieval, visual /
spatial understanding, complexity). When a required check fails, the
annotator is asked to revise the prompt and re-do the
checks.}
\begin{tcblisting}{
    breakable,
    enhanced,
    colback=gray!5,
    colframe=black!70,
    boxrule=0.6pt,
    arc=2pt,
    left=8pt, right=8pt, top=6pt, bottom=6pt,
    title=\textbf{Step 2 of the notebook: Check the prompt against desiderata},
    fonttitle=\bfseries,
    coltitle=white,
    colbacktitle=black!70,
    listing only,
    listing options={
        basicstyle=\ttfamily\footnotesize,
        breaklines=true,
        breakatwhitespace=true,
        columns=fullflexible,
        keepspaces=true,
        showstringspaces=false,
    },
}
For each question below the annotator answers True / False.  If any
*required* check is False, they are sent back to revise the prompt.

=== Utility and Realism (all required) ===
[ ] Task represents something a real user might want to do
    (it has practical value for typical users).
[ ] Task would save users time if automated.

=== Document Synthesis (all required) ===
[ ] Task involves creating or editing a Google Workspace tool
    (Docs, Sheets, Slides, Drive, ...).
[ ] Task requires meaningful synthesis of gathered information
    (not just copying).
[ ] Final product is cohesive and well-organized.

=== Information Retrieval (>= 2 of 3 required) ===
[ ] Task requires retrieving 5+ distinct pieces of information.
[ ] Information must come from 2+ different websites / sources.
[ ] Task explicitly requires browsing the live web.

=== Visual / Spatial Understanding (>= 1 of 3 required) ===
[ ] Task requires identifying and editing specific document elements.
[ ] Task requires spatial-arrangement requirements
    (positioning elements left / right / etc.).
[ ] Task requires color-based instructions or modifications.

  If none are met, the notebook suggests randomized ideas, e.g.:
    "Add web images related to part of the document and place them at a
     given location."
    "(Sheets) Ask user to color-code some elements of the sheet."
    "Ask for certain words to be highlighted some color."
    "Ask to strike out certain words."
    "(Slides) Ask for slides matching some definition to have animation."

=== Complexity (all required) ===
[ ] Task requires 4+ steps for information retrieval
    (search + visit/extract per source).
[ ] Task requires 1+ step to create the document / template.
[ ] Task requires 10+ steps to incorporate information into the document.
[ ] Task cannot be completed by simply pasting information into a
    blank document.

>>  If any required check fails:
    "Maybe go back through previous steps and edit your prompt until it
     meets this requirement. Look at the example above (and in the
     template sheet)."
\end{tcblisting}
\label{fig:colab2}

\captionof{figure}{The checkpoint-writing phase. The notebook shows
detailed instructions and a worked example, then has the annotator write
their own checkpoints in Markdown and answer a final True/False rubric
to verify that each checkpoint is measurable and ties to a concrete
document state change.}
\begin{tcblisting}{
    breakable,
    enhanced,
    colback=gray!5,
    colframe=black!70,
    boxrule=0.6pt,
    arc=2pt,
    left=8pt, right=8pt, top=6pt, bottom=6pt,
    title=\textbf{Step 3 of the notebook: Write and check the checkpoints},
    fonttitle=\bfseries,
    coltitle=white,
    colbacktitle=black!70,
    listing only,
    listing options={
        basicstyle=\ttfamily\footnotesize,
        breaklines=true,
        breakatwhitespace=true,
        columns=fullflexible,
        keepspaces=true,
        showstringspaces=false,
    },
}
=== Instructions for Checkpoints ===
  * Each checkpoint is written in English.
  * Each checkpoint describes a measurable milestone.
  * Checkpoints describe specific state changes in the document.
  * Checkpoints specify one or more required actions:
      - Finding / extracting specific information from websites.
      - Placing / adjusting information in the document in a specific way.
  * For each checkpoint, include:
      - A number
      - A general description
      - Explanation of how to evaluate
      - Eval template(s)

=== Good example (excerpt) ===

  # Checkpoints
  This task has 3 points in total.

  ## Checkpoint 1:
  Check that common information about the letter writer is placed at a
  reasonable location in the document letter.

  ### Outcome Evaluation:
  - Exact match to check that the document includes the writer's right
    name, title, and address.
  - Use OCR to locate the text and ensure it is on the left-hand side of
    the document, beneath the logo.

  ### Eval Template(s)
  - Text Exact Match, Text Location Match

=== Check the checkpoints ===
For each question below the annotator answers True / False.

[ ] Each checkpoint is written in English.
[ ] Each checkpoint describes a measurable milestone.
[ ] Checkpoints describe specific state changes in the document.
[ ] At least one checkpoint requires finding / extracting specific
    information from websites.
[ ] At least one checkpoint requires adding / adjusting information in
    the document in a specific way.
[ ] Task requires 4+ evaluation steps.
\end{tcblisting}
\label{fig:colab3}

\twocolumn

%% file: tables/task_decomposition_example.tex
\onecolumn
\captionof{figure}{Decomposition of one task into checkpoints, and of each
checkpoint into evaluation steps. The example task asks the agent to build
a wedding color-palette spreadsheet; it carries six checkpoints and
twenty-nine evaluation steps. Steps are a mixture of deterministic checks
(URL validity, column structure, exact-text matches) and model-judged
checks (LLM judges for topical relevance, VLM judges for image content
and color-fidelity verification).}\label{fig:task-decomposition}
\begin{tcblisting}{
    breakable,
    enhanced,
    colback=gray!5,
    colframe=black!70,
    boxrule=0.6pt,
    arc=2pt,
    left=8pt, right=8pt, top=6pt, bottom=6pt,
    title=\textbf{Example task with checkpoints and evaluation steps},
    fonttitle=\bfseries,
    coltitle=white,
    colbacktitle=black!70,
    listing only,
    listing options={
        basicstyle=\ttfamily\footnotesize,
        breaklines=true,
        breakatwhitespace=true,
        columns=fullflexible,
        keepspaces=true,
        showstringspaces=false,
    },
}
=== Task Prompt ===

I want to make a Google Sheet to help choose color templates for my wedding. First, do a web search and find at least 3 articles on wedding colors. From those articles, extract at least 10 distinct shades of purple, yellow, or orange. Then, at the top left of the sheet, list all the colors vertically. In the column to the right of the color name, put the link to the article that mentioned the color. Then, in the next column to the right, fill in that cell with a background color matching the named color as closely as possible using the color's standard hex code (e.g. from colorhexa.com). Then, in the next column to the right, find a page or link to that specific color at a paint store and paste the link. Below this list, on the same sheet, add a second table with the same colors as the list above as column headers, in the same order. The rows are different wedding objects with at least 3 objects. Fill in each cell with a real photograph from the internet of that decoration in that color and display it in using =IMAGE() in the sheet. If you can't find an image for an object in a specific color, skip it, but fill in at least half of the cells. Finally, in a new tab, create at least 10 different rows of full-color palettes. For each row, choose three colors from the list above and fill each cell's background color with one of those three colors to show different color palettes.

=== Checkpoints ===

## Checkpoint 1: Color Extraction
The agent found and extracted wedding color information from at least three articles and color names are listed vertically in the top-left area of the main sheet.

### Evaluation Steps:
- All extracted color names appear in a vertical list in a single column.
- The color list starts in the top-left area of the sheet (rows 1-15 approximately).
- At least 10 unique color names/shades are present, extracted from articles.
- The agent searched for and found at least 3 articles about wedding colors (agent trace, LLM judge that the articles are about weddings).
- All extracted colors belong to the specified purple, yellow, or orange categories (LLM judge that each color falls into the category).

## Checkpoint 2: Article Source Links
Article links are placed in the column immediately to the right of each color name.

### Evaluation Steps:
- The article column is a URL/link column (every populated cell contains a URL, not free text).
- Each color name has a corresponding article link in the same row.
- The article links are functional and reachable.
- The links lead to relevant wedding-color content for that color.

## Checkpoint 3: Color Cell Formatting
Cells are filled with colors matching the color names in the third column.

### Evaluation Steps:
- The fill column is a colour-fill column (every populated cell has a background fill, not just typed text).
- The fill colors match the named colors or are extremely close to them (VLM judge).
- The hex value for the cell color matches the color shade hex found in https://www.colorhexa.com/.
- Each color name has a corresponding colored cell.

## Checkpoint 4: Paint Store References
Paint store links are provided in the fourth column next to each color.

### Evaluation Steps:
- The store column is a URL/link column (every populated cell contains a URL, not free text).
- Each color has an associated paint store link.
- The paint store links are functional and reachable.
- The links lead to relevant paint store / color content for that color.

## Checkpoint 5: Wedding Decoration Matrix
A wedding decoration matrix is created below the color list with images.

### Evaluation Steps:
- At least 3 types of wedding decorations are listed in the leftmost column (API for location, LLM judge for content).
- Column headers contain the same color names from the original list (exact text match).
- The matrix column headers appear in the same order as the original color list.
- At least half of the matrix cells contain images.
- Images show the specified decoration type in the corresponding color (VLM judge).

## Checkpoint 6: Color Palette Tab
A separate tab contains at least 10 color palette combinations.

### Evaluation Steps:
- A new sheet/tab was created for color palettes.
- At least 10 rows of color combinations exist.
- Each row contains exactly 3 colored cells representing a palette.
- Colors are filled as background colors (not just text).
- Color combinations use colors from the original extracted list.
\end{tcblisting}

\twocolumn

%% file: tables/task_prompts_table.tex
\onecolumn
\begingroup
\setlength{\LTpre}{0pt}\setlength{\LTpost}{0pt}
\small
\renewcommand{\arraystretch}{1.15}
\begin{longtable}{@{}p{0.22\linewidth} p{0.74\linewidth}@{}}
\caption{Example task prompts from KNOWS. One randomly chosen instance per task; URLs replaced with \texttt{[redacted]}.}\label{tab:task-prompts}\\
\toprule
\textbf{Shortname} & \textbf{Task Instruction} \\
\midrule
\endfirsthead
\multicolumn{2}{c}{\tablename\ \thetable{} -- continued from previous page} \\
\toprule
\textbf{Task} & \textbf{Prompt} \\
\midrule
\endhead
\midrule
\multicolumn{2}{r@{}}{\footnotesize\itshape continued on next page} \\
\endfoot
\bottomrule
\endlastfoot
\texttt{docs\_\allowbreak{}1\_\allowbreak{}formal\_\allowbreak{}letter} & Structure a formal letter written in my name using Google Docs.  \newline Use my website for my info: [redacted]. \newline Include my employers logo in the upper left corner if available online. \newline Include the following information in the letter header: \newline - My Name \newline - My Email Address \newline - My Job/Position Title \newline  \newline Take the photo of my signature from the drive folder here: [redacted] and add it to the end of the letter. \\
\arrayrulecolor{black!20}\hline\arrayrulecolor{black}
\texttt{docs\_\allowbreak{}11\_\allowbreak{}personal\_\allowbreak{}recipe\_\allowbreak{}ocr} & Create a recipe from the Recipe Coral template in Google Docs. First, replace the name of the recipe with the correct recipe name by looking at the image saved in the drive at [redacted]. Then, copy the ingredient list into the template and paste the instructions into the corresponding field. Get tips from an online source, and make sure to cite source urls for tips below the tips entries. Use direct quotes for all tips. Replace the default image with a cropped image from the original drive link that shows the recipe text clearly. Fill out the rest of the template with relevant information from the recipe. Then search the internet for at least 3 similar recipes and put them on separate pages with the same format as the first recipe. Put the link for the source of each additional recipe on a line below the title fitting with the formatting. Make sure each page of recipes is visually distinct from the others. Finally, check the other online recipes and fill in the tips with some useful information that you found. \\
\arrayrulecolor{black!20}\hline\arrayrulecolor{black}
\texttt{docs\_\allowbreak{}31\_\allowbreak{}education\_\allowbreak{}lesson\_\allowbreak{}plan} & I am an Elementary School educator, and I have a free class to teach a fun topic. My class is on Deep Learning. Can you help me make a lesson plan document appropriate for my students? First, come up with 2-3 fun topics in Deep Learning that the class might enjoy but is still related and list those as the main bullets. Then, under each of these, find 4 unique websites that have information about the topic and put their URLs as sub-bullets. Then, for each of these websites, list 5 facts as sub-bullets to the url. Finally, list all of the facts below the bullets and color code them by which topic they are about with the highlighter. Use a different color for each topic. Write a summary of the lesson using the facts collected. Finally, go online to find images related to each of the topics. Put them all side-by-side at the bottom of the last page of the document. \\
\arrayrulecolor{black!20}\hline\arrayrulecolor{black}
\texttt{docs\_\allowbreak{}37\_\allowbreak{}reference\_\allowbreak{}list} & Create a Google Doc reference list from this course webpage: [redacted]. For each lecture, add a Heading 3 with 'Module N: Lecture title', where N is the module number. Things like exams and holidays should not appear in the reference list. Under each title, create a separate bullet list for tutorials, textbooks, and videos. Each bullet list should start with "Tutorials", "Textbooks", or "Videos" in bold, normal font. If a given type has no hyperlinks, delete that category. Each bullet is a hyperlink to a resource referenced in the lecture slides, using the resource title as the anchor text instead of the raw URL. Include author(s) in parentheses after the hyperlink, formatted as "(First author)". Next to the hyperlink, add "Slide: <list of slide numbers separated by commas>" listing the slides where the link is referenced. No duplicates: if something is referenced multiple times, then add all slide numbers. If a link is referenced on more than one slide, bold that hyperlink (and only the hyperlink) in the corresponding bullet list, italicize it, color it 'dark cyan 1', and set its font size to 12pt. If the link is inactive (dead link) do not include it in the google doc. \\
\arrayrulecolor{black!20}\hline\arrayrulecolor{black}
\texttt{docs\_\allowbreak{}5\_\allowbreak{}influential\_\allowbreak{}papers} & Create a Google Docs with five influential papers on parameter-efficient fine-tuning (PEFT). Include links to arxiv.org for each of the papers. The papers should have at least 100 citations, have been first published within the last three years, and must have an active page on semanticscholar.org. Include the name of the paper, a link below the name, and the abstract of each of the papers below the link. \\
\arrayrulecolor{black!20}\hline\arrayrulecolor{black}
\texttt{sheets\_\allowbreak{}10\_\allowbreak{}paper\_\allowbreak{}sorting} & Create a new google sheets document for all the papers in my drive folder [redacted]. Make a blank google sheets table and for each paper, in column A put the title of the paper, in column B put the paper authors, in column C put the abstract, in column D put a link to the arxiv abstract website, in column E put the links to the google drive pdfs. Additionally, search the web for at least three more papers on arxiv where the first author of each existing paper is also listed as the first author (or less if they do not have at least 3 first-authorships on arxiv) and put the same information. For new papers, download the PDFs and upload them to the pdfs folder in this drive folder: [redacted] and link them in column E. Extract or take a screenshot of Figure 1 for each paper, upload the image to the figures folder in this drive folder: [redacted] and display it in column F using =IMAGE() with the Drive URL. In column G put a checkbox and check for each paper that is a new paper. Highlight papers that have a related works sections which mention dark energy in yellow, and make sure to group these papers at the top rows of the sheet. Ensure all of the text in the table is visible and not cut off. \\
\texttt{sheets\_\allowbreak{}2\_\allowbreak{}personal\_\allowbreak{}recipe\_\allowbreak{}foodcomposition} & From the recipe [redacted], pull the average amount of iron, carbohydrates, fat, sugar, potassium, vitamin A, protein, sodium, fiber, calcium, and vitamin C for each ingredient (based on its quantity) using the USDA food database. When searching the USDA, use the food code that best matches the ingredient; more general codes are usually preferable, unless the recipe specifies a matching brand. For instance, if a listed ingredient is just ``Paneer'', you would use ``Cheese, paneer'' instead of ``Palak Paneer''. You may use any of the five USDA food search tabs. Place the results into a Google Sheet as follows. Group nutrients into macros, minerals, and vitamins. Within each group, sort nutrients alphabetically. The first column should be titled "Ingredients" and the second ``Link'', followed by sorted nutrient columns starting with macros, then minerals, then vitamins. Above each group, add a merged, centered, italicized header cell labeled "Macros", "Minerals", or "Vitamins" spanning all associated columns. Bold all column titles, excluding the group header cells. Color all cells in each group with a light background color, making sure each group has a distinct color. Place a horizontal line under column titles. List ingredients in the first column in the order they appear in the recipe. In the ``Link'' column, add the USDA data link you used to obtain the quantity of a nutrient in a specific food. Bold any nutrient values exceeding 10\% DV according to FDA guidelines. For ingredients you can't find in the USDA database, leave centered N/A. For ingredients with a range of values as the amount, use the minimum value. Ignore ingredients that are listed as optional or don't have a specified amount. \\
\arrayrulecolor{black!20}\hline\arrayrulecolor{black}
\texttt{sheets\_\allowbreak{}28\_\allowbreak{}personal\_\allowbreak{}travel\_\allowbreak{}planner} & A family of 4 people is traveling to Washington D.C this May for 3 days, and they need a detailed day-by-day itinerary. Please prepare this travel plan and put everything into a Google Sheet so it's easy to follow. They will stay at the same hotel Hyatt Regency Washington on Capitol Hill for all 3 days of the trip and use Public Transportation for travel during the trip. They plan to do a weekday trip. \newline  \newline Here's what they are looking for: \newline - Activities \& Sightseeing: Each day should have a mix of morning, afternoon, and evening plans at popular attractions in the city. Please include the opening hours, suggested duration, and any entrance fees. (For Activities \& Sightseeing, there should be three rows per day: morning, afternoon, evening.) \newline - Food Stops: Include two food stops per day (lunch and dinner). For each, add the type of cuisine, an average price estimate. \newline - The food stops must be in the order: Lunch between Morning and Afternoon activity, Dinner between Afternoon and Evening activity. \newline - Time of Day: Each row must specify the time slot --- one of: Morning, Afternoon, Evening (for activities), or Lunch, Dinner (for food stops). \newline - Start Time and Departure Time: Each row must include a start time (when the visit/meal begins) and a departure time (when the family leaves). These times must align with the Time of Day slot (e.g., Morning activities start in the morning hours, Lunch starts around midday). Start and departure times must not overlap with adjacent events and must leave enough time for transportation between destinations. The schedule across each day should flow sequentially without gaps or conflicts. \newline - Transportation: For every entry (activity or food), note how they'd get between places (walking, metro, bus, taxi, etc.) and the typical travel time. Clearly indicate whether each day is a weekday or weekend, and adjust transit time estimates accordingly (e.g., account for rush hour delays on weekdays and reduced service or increased congestion on weekends). Ensure transit time correctness reflects these differences. \newline - Costs: Provide an estimated cost for each main activity and meal, and color-code the cost column: \newline under \$100 = green, \$100--\$200 = yellow, above \$200 = orange \newline - The review link for both Activity and Food should be from Google Maps. \newline - Alternatives: For every destination (food or activity), suggest one backup option (for weather, crowds, or unexpected closures). Put these in the last column and format them in blue. \newline - Please avoid repeating the same destination or restaurant across the trip. \newline - If two places are far apart, adjust the order so the day makes geographic sense, and the pacing feels realistic (not jam-packed, but no big idle gaps). \newline - Include the following columns: Date, Time of Day, Destination (Food/Activity), Cuisine, Opening time, Start time, Departure Time, Duration, Review link, Transportation Mode, Travel Time, Cost (with color coding), Alternative Option (in blue). \newline - Freeze the top row (header). \newline - Individual transit times should be <= 30 minutes and keep total transit times under 90 min per day \\
\texttt{sheets\_\allowbreak{}25\_\allowbreak{}skitourplan} & Create a ski tour plan for 3 possible runs 32 degrees or less using the wasatch backcountry ski guide. Organize the route information for each run in a google sheet including the run name, run link, starting location, gps coordinates of the run location, typical vertical of the run, slope aspect, and slope angle as columns. Then look up the danger color rating for this run according to the Utah Avalanche Center forecast for 12/28/2024. Add columns for forecast date and forecast link and a screenshot of the full danger rose. These 3 columns should be merged vertically since they apply to all runs. Then using this information, determine the danger color rating for each run. Color the cell of each run name with the color of its danger rating. \\
\arrayrulecolor{black!20}\hline\arrayrulecolor{black}
\texttt{sheets\_\allowbreak{}38\_\allowbreak{}apartment\_\allowbreak{}finder} & I am looking to rent a 3 bed 2 bath unit in Chicago, Illinois. Your task is to check www.craigslist.org and create a google sheet with at least 6 listings. My budget is 3000-4500 USD/month and I would like to move in within the next month. The unit should have the features: air-conditioning and optionally has at least 1200 sq ft. For each unit include the necessary columns (including the address) and a link to the listing in the spreadsheet and use conditional formatting on the numeric features (green to red where green means better and red means worse). Add an additional column at the end on interesting positive features of the listing and another column on potential dealbreakers (if there are none, keep it empty). Create a separate "summary statistics" table in the top-right corner (starting at column L) while the main data starts at A1. Populate the summary statistics table with data about the listings table using sheets equations that will update the statistics automatically. Ensure all of the text in both tables is visible and not cut off. Both tables must be inserted as blank Google Sheets Tables and then populated with the data. \\
\arrayrulecolor{black!20}\hline\arrayrulecolor{black}
\texttt{sheets\_\allowbreak{}45\_\allowbreak{}Personal\_\allowbreak{}WeddingPlanner\_\allowbreak{}weddingcolorpallette} & I want to make a Google Sheet to help choose color templates for my fall wedding. First, do a web search and find at least three articles discussing fall or autumn wedding colors. From those articles, extract at least 15 different specific shades of burgundy, rust, or mustard. Then, at the top left of the sheet, list all the colors vertically. In the column to the right of the color name, put the link to the article that mentioned the color. Then, in the next column to the right, fill in that cell with a background color matching the named color as closely as possible using the color's standard hex code (e.g. from colorhexa.com). Then, in the next column to the right, find a page or link to that specific color at a paint store and paste the link. Then, in the next column, add a short one-sentence description of the mood or feeling that color evokes. Below this list, on the same sheet, add a second table with the same colors as the list above as column headers, in the same order. The rows are different wedding objects with at least 4 objects. Fill in each cell with a real photograph from the internet of that decoration in that color and display it in using =IMAGE() in the sheet. If you can't find an image for an object in a specific color, skip it, but fill in at least half of the cells. Finally, in a new tab, create at least 15 different rows of full-color palettes. For each row, choose three colors from the list above and fill each cell's background color with one of those three colors to show different color palettes. Add a label in the first column of each row giving the palette a creative name. \\
\arrayrulecolor{black!20}\hline\arrayrulecolor{black}
\texttt{sheets\_\allowbreak{}55\_\allowbreak{}Movie\_\allowbreak{}Recommendation} & I want to watch a movie tonight. I prefer the following genres: Action, Drama, Thriller, Sci-Fi, and Comedy. Find the genre based on the IMDb listing of the movies. \newline I want to watch award-winning movies, and I prefer movies that are highly rated and well-reviewed (IMDb $\geq$ 6.5). Look for movies that have won at least one Oscar award for Best Actor, Best Actress, Best Director, Best Original Screenplay, Best Adapted Screenplay, or Best Cinematography. \newline Please make a Google Sheet with at least 5 movie recommendations that won at least one of the mentioned awards. Each movie's primary genre on IMDb --- the first standard genre IMDb lists for the movie --- must be one of the above listed genres. Use IMDb's standard genre list, not UI tags. \newline Each row should contain information about one movie. Use these column header names, in this order: \newline Movie Title \newline Genre \newline MPA/Age Rating \newline IMDb Score \newline Release Year \newline Duration \newline Oscar Awards Won \newline  \newline MPA/Age Rating is the MPA/age rating, e.g., PG, PG-13, R. \newline  \newline In the Oscar Awards Won column, list all of the qualifying Oscar categories above that the movie won, separated by commas. Do not include other awards. \newline Sort the list based on Duration in descending order. \newline Highlight the highest IMDb score in the list with green fill + bold text. \newline Highlight the lowest IMDb score with red fill + bold text. \\
\texttt{sheets\_\allowbreak{}6\_\allowbreak{}investmenttracker} & Make a stock tracker google sheet for the top 10 highest market cap stocks in tech from the end of Q2 2023. Insert a blank Google Sheets Table and then populate it with the data. Add the name and ticker of each of these to my watchlist and compare their past price to the current price for each stock to determine how much I would have gained or lost on each stock. Finally, put that I currently own 100 shares of each of these stocks and make a bar chart showing each of my stocks and their total value as a percentage of the overall portfolio. \\
\arrayrulecolor{black!20}\hline\arrayrulecolor{black}
\texttt{sheets\_\allowbreak{}7\_\allowbreak{}running\_\allowbreak{}analysis} & Gather my running data from the provided Drive link: [redacted] and put relevant data into a table in a Google Sheet. Create 2 plots using this data, one that shows my average running speed in (min/mile) over time, and one that shows the cumulative distance I've ran in miles over time. The running speed plot should have all runs as circular points and the distance plot should be a tracking the increasing distance over time. Include these plots in the same sheet tab as the data table. Find online and add in the average male age 25 running speed for a 5k as a baseline dotted line in the average running speed plot. Compare this to the average running speed for Eliud Kipchoge in his top 3 races in the Marathon. Make sure everything in the google sheet is fully unobscured and put the sources for the average running times you retrieve below the relevant plot. \\
\arrayrulecolor{black!20}\hline\arrayrulecolor{black}
\texttt{slides\_\allowbreak{}17\_\allowbreak{}removeimagesaddplacehold\-ers} & Take the presentation at [redacted] and help me replace all the images. First, make a copy of the original presentation and put in into [redacted]. Then identify all of the images in the document and save them all as image files and place them into [redacted] for possible later use. Delete all the images, replacing each one with a description of the image in a text box at the same spot in big red text. Finally, for each of the images, find a similar image online and place it in the right spot. For each of these, make sure to put the URL for the image address below each new image to credit it. \\
\arrayrulecolor{black!20}\hline\arrayrulecolor{black}
\texttt{slides\_\allowbreak{}20\_\allowbreak{}Illustrated\_\allowbreak{}Book\_\allowbreak{}Report} & I need to make a book report slideshow about A Cat At the End of the World, can you help me get started? First, on the title slide put my name Anisija Mihaljevi\'c on the bottom, the title of the book above my name, and a photo of the cover of the book above that. Next, make a slide for at least 5 of the major characters with their name as the title and bulletpoints describing their characteristics below. At the bottom of each slide, include a section where you link to the source that you got the information for that character from. Make sure each character slide has a different background color. Finally, make a slide for the author of the book with their name and a photo of them. At the end, create a references slide and include links to at least 3 different sites that you used to create the presentations. Ensure that all information you enter in the slides is from these references. \\
\arrayrulecolor{black!20}\hline\arrayrulecolor{black}
\texttt{slides\_\allowbreak{}26\_\allowbreak{}basic\_\allowbreak{}educational\_\allowbreak{}slide\_\allowbreak{}deck} & I'm preparing an introductory teaching presentation on 'Mercury' for a 7th-grade audience, and I'd like help designing the slide content from scratch. \newline  \newline Start with a title slide that highlights the Mercury in large, bold, dark-orange lettering, includes my name (Mrs. Emily Carter), and uses a thematically relevant background image. Add a small-font source credit beneath the image. \newline  \newline Next, break the Mercury into four student-friendly sections. For each section, locate one reliable online reference through web search, and create a slide that includes: \newline - A bold-italic section heading \newline - A short list of bullet-point explanations based on the source you found \newline - A clear citation for the source in small type in the lower-left corner \newline - An illustrative online image, placed on the right side, with its source credit beneath it in small text. \newline  \newline Finish with a slide that summarizes the key ideas and encourages the student to ask questions. \\
\arrayrulecolor{black!20}\hline\arrayrulecolor{black}
\texttt{slides\_\allowbreak{}29\_\allowbreak{}buy\_\allowbreak{}car\_\allowbreak{}pres} & Create a slides presentation comparing different cars that can be categorized as a sports car. First, find an online article talking about at least 5 different 2010 models. Make a title slide called "Comparing Different Cool Cars to Buy" with font size atleast 30pt. Then for each of at least 5 cars of that category, make a slide for each. Start with the make and model as the slide title. Then, for each car, go to Kelly Blue Book and find the sticker price, fuel efficiency and horsepower. Then, go to Kelly Blue Book for user reviews and get the average rating for each. Include the link to the review website in the slide. For each car, add a picture of the car that takes up at least 50\% of the slide. The picture must be from the article or Kelly Blue Book. Finally, on the last slide, list which car had the best stats in each category (sticker price, fuel efficiency, horsepower, user rating). \\
\texttt{slides\_\allowbreak{}30\_\allowbreak{}Work\_\allowbreak{}Wikipedia\_\allowbreak{}Photos} & I represent several clients in Science and want to help replace their bad Wikipedia pages. I need to make a presentation on this. Could you please look at my client list at [redacted] and make a presentation for me? The title should be "Why we need new Wikipedia headshots" and have a picture on the top right taking the majority of the slide of Tom Hanks from their wikipedia page. Then, for each of my clients, make a slide with their name at the top, their current wikipedia picture on the left. Then, find images on the internet of each client and put it on the right side so it is symmetric with their wikipedia picture. Finally, on the last slide, copy all of the wikipedia urls of the client pages and then the webpage of a photographer in San Francisco. \\
\arrayrulecolor{black!20}\hline\arrayrulecolor{black}
\texttt{slides\_\allowbreak{}39\_\allowbreak{}Personal\_\allowbreak{}Lookbook\_\allowbreak{}PaintColors} & Create a Google slides presentation to help us choose colors for our home office. First, on the title slide, find an image of a home office by searching "home office" and choosing one of those images. Make home office the title text of the title slide and make the image as big as possible. Then, for the next 5-10 slides, choose a color. Make the color the title of each of those slides. Then for each slide, search for images of that color and home office and choose images with that color. Put two of these images on the slide, one on the bottom left and one on the bottom right. Ensure that each image's source URL is present in the image's ALT text. Finally, on the final slide, choose whichever color from the previous slides you like best. Then on that final slide write in big font "COLOR is the best choice" and make the background color of only that slide that same color. \\
\arrayrulecolor{black!20}\hline\arrayrulecolor{black}
\texttt{slides\_\allowbreak{}51\_\allowbreak{}event\_\allowbreak{}announcement\_\allowbreak{}poster} & Create a one-page Google Slides formatted as a visually appealing event announcement poster. This is the Breaking the Silence: Understanding the Youth Mental Health Crisis event, organized by the American Psychological Association. It will take place on 10/03/2026 at the Ronald Reagan Building and International Trade Center, Washington, D.C. The main theme is the growing mental health crisis among adolescents and young adults, and the featured speaker is Dr. Lisa Damour. \newline Retrieve reliable background information about the growing mental health crisis among adolescents and young adults that is closely related to Dr. Lisa Damour's work from a few stable, reputable sources (e.g., universities, professional associations, encyclopedias, major news outlets, and the speaker's website). Summarize the topic in 1-2 paragraphs that highlight why the event is worth attending and cite the websites you used. Also, introduce the speaker, including their education and current affiliation, and mention 3 of their main professional achievements. \newline Poster Structure: \newline Header (top, centered): Breaking the Silence: Understanding the Youth Mental Health Crisis, large, bold, in color Deep Teal. \newline Subheader (below header, left-aligned): 10/03/2026, Ronald Reagan Building and International Trade Center, Washington, D.C., and hosted by the American Psychological Association, styled in a smaller italic font. \newline Logo (top, right): Retrieve the logo of the American Psychological Association and insert it to the top right of the poster. Make sure the logo size isn't too big or too small. \newline Central Section (main body, centered): Synthesized background on the growing mental health crisis among adolescents and young adults (engaging, persuasive, 1-2 short paragraphs). \newline Right Sidebar (shaded box, color Warm Amber): Speaker bio (Dr. Lisa Damour), including her credentials, achievements, and relevance to youth mental health. \newline Footer (bottom, right-aligned): Contact info of the American Psychological Association, Dr. Lisa Damour's webpage. \newline Styling Requirements: Ensure visual contrast between header, subheader, and body text. Use at least two colors: Deep Teal for header elements and Warm Amber for sidebars/highlights. Maintain consistent spacing and alignment so the document reads like a poster, not just plain text. Ensure that citations are placed in speaker notes and say exactly what they are citing in the poster. \\
\texttt{slides\_\allowbreak{}42\_\allowbreak{}personal\_\allowbreak{}none\_\allowbreak{}product\_\allowbreak{}comparison} & My colleague, James, has just been promoted to Senior Account Manager and will be starting his new role at Deloitte next quarter. I want to buy him a tablet as a congratulatory gift and need help deciding between three options: iPad Pro (M4), Samsung Galaxy Tab S10 Ultra, and Microsoft Surface Pro 11. \newline Objective Create a comprehensive, visually appealing slide deck that compares these three devices specifically for a corporate professional's use. The presentation should be polished and persuasive rather than dry or overly technical, helping both my colleague and me make an informed decision. \newline Target Audience Primary: Newly promoted corporate professional entering a client-facing role Secondary: Colleague making the purchase decision \newline Design Requirements Theme: Use Deloitte's official colors (Deloitte Green and black) throughout the presentation Tone: Professional, confident, and easy to understand Focus: Business professional needs and workplace lifestyle \newline Slide Structure and Content \newline Slide 1: Title Slide \newline Title (bold): "A Gift for James!" \newline Subtitle: "A Comparison of the iPad Pro (M4), Samsung Galaxy Tab S10 Ultra, and Microsoft Surface Pro 11" \newline Image: Photo representing Deloitte (office or branding scene) \newline The title and subtitle are to the left, and the image is to the right of the slide \newline Slide 2: The Challenge \& The Goal \newline Explain the decision-making challenge \newline Define what we're trying to achieve \newline Slide 3: How We'll Judge (The Criteria) \newline List evaluation criteria relevant to business professionals: \newline Client presentation capability \newline Battery life for long workdays and travel \newline Portability and weight for commuting \newline Performance for productivity and multitasking \newline Security and enterprise compatibility \newline Other relevant factors \newline Slides 4-6: Deep Dive on Each Device \newline One slide per device (iPad Pro M4, Samsung Galaxy Tab S10 Ultra, Microsoft Surface Pro 11) Content per slide: \newline Key features and specifications \newline Pros and cons with respect to professional/corporate use \newline Two product images showing different angles \newline All the information should come from the official website of each option. Cite the resources when needed. \newline Slide 7: Side-by-Side Comparison \newline Format: Three-column comparison table, with devices as column headers \newline Content: Feature-by-feature comparison using the established criteria \newline Color coding system: \newline Green: Best performer in each category \newline Yellow: Middle performer \newline Red: Lowest performer in that category \newline Slide 8: Who Is Each Device For? (The Recommendation) \newline Summarize information on each option and match each device to different professional types/needs \newline Provide clear recommendations based on different priorities \newline Additional Guidelines \newline Keep technical jargon to a minimum \newline Use business-relatable examples and scenarios \newline Include practical considerations for corporate life (travel, client meetings, remote work) \newline Ensure all information is current and accurate \newline Make the presentation visually engaging with appropriate use of images, colors, and formatting \\
\end{longtable}
\endgroup
\twocolumn

%% file: prompts/claude_code_prompt.tex
\onecolumn

\captionof{figure}{The prompt we use with Claude Code to draft an evaluator script for a given task. The generated scripts were a useful starting point, but finalizing them and making them reliable took substantial effort of multiple authors.}\label{fig:claude-code-prompt}

\begin{tcblisting}{
    breakable,
    enhanced,
    colback=gray!5,
    colframe=black!70,
    boxrule=0.6pt,
    arc=2pt,
    left=8pt, right=8pt, top=6pt, bottom=6pt,
    title=\textbf{Prompt for drafting evaluator scripts with Claude Code},
    fonttitle=\bfseries,
    coltitle=white,
    colbacktitle=black!70,
    listing only,
    listing options={
        basicstyle=\ttfamily\footnotesize,
        breaklines=true,
        breakatwhitespace=true,
        columns=fullflexible,
        keepspaces=true,
        showstringspaces=false,
    },
}
# Task Creation Guidelines

> **IMPORTANT: Before writing or modifying any evaluator, read [`eval_utils/EVAL_UTILS_DIRECTORY.md`](../eval_utils/EVAL_UTILS_DIRECTORY.md) in full.** It is a comprehensive directory of every shared utility function with descriptions and real usage examples. Always reuse existing utilities from `eval_utils/` rather than writing new helper functions. This is essential for maintaining consistency across evaluators and avoiding duplicate code.

## Directory Structure

tasks/
+-- <template_name>/
    +-- utils.py              # Shared utilities (optional)
    +-- DEV.md                # Development state tracking (required when developing)
    +-- test/                 # Test suite (optional)
    +-- instance_1/
        +-- task.md           # Task description
        +-- checkpoints.md    # Evaluation rubric
        +-- evaluator.py      # Evaluation logic
        +-- id.txt            # Unique identifier
        +-- gold_instances.csv
        +-- data/             # Task-specific assets (optional)

### Required Files (Instance Level)
- **`task.md`**: Human-readable task description
- **`checkpoints.md`**: Evaluation criteria with point values
- **`evaluator.py`**: Implements `grade_checkpoints()` returning a `Result` object
- **`id.txt`**: Unique task identifier
- **`gold_instances.csv`**: Maps instance names to Google document IDs

### Optional Files
- **`utils.py`** (template level): Shared utilities - check `eval_utils/` first before creating new ones
- **`data/`**: Reference datasets, gold images, expected outputs
- **`test/`**: Unit and integration tests

---

## DEV.md - Development State Tracking

**CRITICAL**: When developing or modifying an evaluator, maintain a `DEV.md` file at the template level. This file tracks development state, issues, and decisions.

### Requirements
- **Maximum 300 lines** - keep it concise and relevant
- **Read DEV.md** at the start of each session working on the evaluator
- **Update DEV.md** after significant changes or when issues are discussed
- **Delete irrelevant sections** as issues are resolved

### DEV.md Structure
# <Task Name> - Development Notes

## Current Status
Brief summary of evaluator state (working/in-progress/blocked)

## Known Issues
- Issue 1: description and status
- Issue 2: description and status

## Recent Changes
- YYYY-MM-DD: What changed and why

## Implementation Decisions
Key decisions made and their rationale

## TODO
- [ ] Remaining work items

## Notes from Discussions
Relevant context from user conversations

### When to Update DEV.md
- After implementing a checkpoint
- When encountering a bug or edge case
- After discussing issues with the user
- When making non-obvious implementation decisions
- Before ending a development session

---

## Checkpoint Mapping

**Each criterion in `checkpoints.md` maps to exactly ONE `add_step()` call.**

Exception: Repeated validations (e.g., 5 slides) use loops with clear naming like `f"Slide {i+1} - Title"`.

### Correct Example
# checkpoints.md
## Checkpoint 1 (3pt): Title slide elements
- Student name present
- Book title present
- Cover image present

def grade_checkpoint_1():
    checkpoint = Checkpoint(total=3, result=0, name="Title Slide")
    checkpoint.add_step("Student Name", check_name(), 1, "...")
    checkpoint.add_step("Book Title", check_title(), 2, "...")
    checkpoint.add_step("Book Cover", check_cover(), 3, "...")
    return checkpoint

### Incorrect - Don't Split Criteria
# DON'T: One criterion becoming multiple steps
checkpoint.add_step("Bullet Count", has_bullets, 1, "...")
checkpoint.add_step("Bullet Content", valid_content, 2, "...")  # Combine these!

---

## Planning Evaluation Steps

For each step, document:

1. **Summary**: 1-2 sentence implementation description
2. **Utilities**: List from `eval_utils/`, `utils.py`, or other evaluators
3. **Hierarchy**: Fallback chain (exact -> fuzzy -> LLM-based)
4. **New Utilities**: If needed, specify name, purpose, and location

Example:
### Step 1.2: Book title (fuzzy match)
**Summary**: Validate title appears on slide with minor variation tolerance.
**Utilities**: `text_fuzzy_match_contained()`, `get_slide_text_content()`
**Hierarchy**: exact match -> fuzzy (85%
**New Utilities**: None

---

## Code Patterns

### Imports
import os, time
from typing import List

from rapidfuzz import fuzz
from src.browsergym.eval.eval_utils.scoring import Checkpoint, Result
from src.browsergym.eval.eval_utils.text_utils import (
    keyword_exact_match,
    keywords_exact_match,
    keywords_match_robust,
)

### LLM Queries
from src.browsergym.eval.eval_utils.models import load_model

model = load_model("gemma-google-ai")
messages = [
    {"role": "system", "content": [{"type": "text", "text": "Instructions"}]},
    {"role": "user", "content": [
        {"type": "image", "image": "/path/to/image.png"},  # For vision
        {"type": "text", "text": "Query"}
    ]}
]
response = model(messages)

### Evaluator Structure
def grade_checkpoints(workspace_doc_id, cached_models=None, browsing_history=None):
    total_start = time.time()
    checkpoints = [grade_checkpoint_1(), grade_checkpoint_2()]
    return Result(checkpoints, total_execution_time=time.time() - total_start)

def grade_checkpoint_1():
    start = time.time()
    checkpoint = Checkpoint(total=5, result=0, name="Checkpoint Name")

    step_start = time.time()
    success = perform_validation()
    checkpoint.add_step("Step Name", success, 1, "Details", time.time() - step_start)

    checkpoint.execution_time = time.time() - start
    return checkpoint

---

## Best Practices

1. **One criterion = one step** (except loops)
2. **Reuse before create** - check `eval_utils/` first
3. **Track timing** for performance monitoring
4. **Specific exceptions** - no bare `except`
5. **Detailed failure messages** for debugging
6. **Clean up temp files** after evaluation
7. **Support cached models** to avoid reloading

## Common Pitfalls

- [X] Multiple steps for one criterion
- [X] Bare `except` statements
- [X] Missing failure details
- [X] Hardcoded paths
- [X] Duplicating existing utilities
- [X] Forgetting to update DEV.md

---

## Rules

**These rules are mandatory and must always be followed.**

1. **No utility methods in `evaluator.py`**: Never define helper functions or utility methods directly in evaluator files. All utilities must go in:
   - `eval_utils/` - for functions reusable across multiple tasks
   - `utils.py` (template level) - for task-specific shared utilities

   Evaluators should only contain `grade_checkpoints()` and `grade_checkpoint_N()` functions.

2. **Run eval-utils-reviewer agent after major evaluator changes**: After making significant updates to an evaluator (adding checkpoints, modifying evaluation logic, refactoring), run the `eval-utils-reviewer` agent to verify compliance with utility placement rules and identify any helper functions that should be moved.

---

## Running Evaluators

### Prerequisites
source <path-to-venv>/bin/activate

### Via VSCode (Recommended)
Check `.vscode/launch.json` for debug profiles with required args.

### Command Line
python src/browsergym/eval/tasks/<template>/instance_1/evaluator.py --workspace_doc_id="<id>"

---

## Checklist

- [ ] `task.md` describes the task clearly
- [ ] `checkpoints.md` lists criteria with points
- [ ] Each criterion maps to one step in `evaluator.py`
- [ ] `id.txt` contains unique identifier
- [ ] `DEV.md` tracks development state (when developing)
- [ ] Utilities tested and documented
- [ ] Error handling is robust
- [ ] Temp files cleaned up
\end{tcblisting}

\twocolumn

%% file: prompts/step_categorization.tex
\onecolumn
\captionof{figure}{The prompt we used to produce categorization in Table \ref{tab:eval_templates}.}
\label{listing:cat_prompt}
\begin{tcblisting}{
    breakable,
    enhanced,
    colback=gray!5,
    colframe=black!70,
    boxrule=0.6pt,
    arc=2pt,
    left=8pt, right=8pt, top=6pt, bottom=6pt,
    title=\textbf{Categorization of Evaluation Step Check},
    fonttitle=\bfseries,
    coltitle=white,
    colbacktitle=black!70,
    listing only,
    listing options={
        basicstyle=\ttfamily\footnotesize,
        breaklines=true,
        breakatwhitespace=true,
        columns=fullflexible,
        keepspaces=true,
        showstringspaces=false,
    },
}
# Eval Step Taxonomy Agent Prompt

You are categorizing every evaluation step across a benchmark dataset of Google Workspace tasks. Your goal is to produce one JSON file per task instance that classifies each eval step into a taxonomy category and flags whether the step primarily uses an LLM/VLM for evaluation.

## Task Overview

The benchmark lives under `src/browsergym/knows/eval/tasks/`. There are 22 tasks, each with 5 instances (`instance_1` through `instance_5`). For each instance you will:

1. Read **three files** to understand what the eval step checks and how it is implemented:
   - `checkpoints.md` -- describes what each step evaluates in plain language
   - `evaluator.py` -- implements the evaluation logic (shows which utility functions are called)
   - Task-level `utils.py` (one directory up from the instance, e.g. `src/browsergym/knows/eval/tasks/<task_name>/utils.py`) -- shared helpers called by the evaluator. This file may not exist for every task; skip if absent.

2. For each eval step listed in `checkpoints.md`, assign exactly one **category** from the taxonomy below.

3. For each eval step, determine whether an **LLM or VLM is the primary evaluation mechanism** (see rules below).

4. If a step does not fit cleanly into one category, **flag it** for human review.

5. Write a JSON output file to `output/taxonomy/<task_name>__<instance>.json`.

---

## Taxonomy Categories

### 1. `visual_check`
The step requires a Vision-Language Model (VLM) as its primary check, or compares an image against another image (pixel-level, perceptual hash, or template matching).

**Examples from the benchmark:**
- "The wiki image of Tom Hanks is present on the slide." (slides_30, checkpoint 1)
- "An image from the client's Wikipedia page is found on the left side of the slide." (slides_30, checkpoint 2)
- "Logo is the University of Utah's logo" (slides_51, checkpoint 2)
- "Figure 1 images match gold data" (sheets_10)

**Implementation signals in evaluator.py:**
- Calls to `binary_judge_image()`, `image_exact_match()`, `match_image_tiered()`, `perceptual_hash_match()`, `verify_image_in_region()`
- Direct VLM model calls asking "Does this image show X?"

### 2. `structural_check`
The step verifies the existence, order, or hierarchy of elements -- not their styling or position, but whether the right pieces are present and in the right arrangement.

**Examples from the benchmark:**
- "There is a column/row with the name of each stock." (sheets_6, checkpoint 1)
- "'Ingredients' column exists and is first." (sheets_2, checkpoint 1)
- "Macros columns are in alphabetical order." (sheets_2, checkpoint 1)
- "A bar chart was created in the spreadsheet." (sheets_6, checkpoint 4)
- "There are no empty bullet lists." (docs_37, checkpoint 2)
- "No duplicate reference links." (docs_37, checkpoint 3)
- "Subheader exists below the main header" (slides_51, checkpoint 1)

**Implementation signals in evaluator.py:**
- Column existence checks, row ordering logic
- Checks for presence/absence of elements (charts, headers, bullet lists)
- `match_columns()` calls (checking if columns exist)

### 3. `formatting_check`
The step verifies visual styling properties: bold, italic, font size, cell colors, bullet points, highlight colors, text alignment, merged cells, text overflow.

**Examples from the benchmark:**
- "The title of each lecture should be of size 'Heading 3'" (docs_37, checkpoint 1)
- "Group headers are centered and italicized." (sheets_2, checkpoint 2)
- "Column titles are bolded (excluding group headers)." (sheets_2, checkpoint 2)
- "Hyperlinks of references that appear in multiple slides are colored 'dark green 2'" (docs_37, checkpoint 4)
- "Event name text has bold font" (slides_51, checkpoint 1)
- "Event name text has font size >= 20" (slides_51, checkpoint 1)
- "Event name text color is black" (slides_51, checkpoint 1)
- "Macro columns share same background color." (sheets_2, checkpoint 3)
- "Values exceeding 10%
- "A shaded right-hand sidebar in grey exists" (slides_51, checkpoint 4)
- "Horizontal line under column titles exists." (sheets_2, checkpoint 2)

**Implementation signals in evaluator.py:**
- `is_cell_bold()`, `is_cell_italic()`, `is_cell_centered()`
- `get_cell_background_color()`, `colors_are_similar()`, `colors_are_distinct()`
- `get_text_style_from_shape()`, font size comparisons
- `get_paragraph_alignment()`

### 4. `spatial_check`
The step validates the physical position or layout of an element on the page/slide/document -- where something appears, how large it is, or its spatial relationship to other elements.

**Examples from the benchmark:**
- "The Tom Hanks image is positioned in the top right area of the slide." (slides_30, checkpoint 1)
- "The Tom Hanks image takes up more than 50%
- "The Wikipedia image and the internet image are approximately symmetric in vertical position." (slides_30, checkpoint 2)
- "Logo is inserted in the top right of the slide" (slides_51, checkpoint 2)
- "Logo does not overlap or crowd header/subheader text" (slides_51, checkpoint 2)
- "Footer is positioned at the bottom right of the poster" (slides_51, checkpoint 5)
- "Event name text box appears at the top of the page" (slides_51, checkpoint 1)
- "No text or images are overlapping each other" (slides_51, checkpoint 7)
- "No text or images are off of the slide" (slides_51, checkpoint 7)
- "Background summary is placed in the main, central body of the poster" (slides_51, checkpoint 3)

**Implementation signals in evaluator.py:**
- `is_text_in_title_position()`, `get_element_bbox()`, `get_slide_dimensions()`
- `is_upper_left()`, `is_upper_right()`, `is_mostly_inside()`
- `get_image_area_percentage_from_api()`
- Bounding box arithmetic, overlap detection

### 5. `content_check`
The step verifies that information came from the correct source, or that content is a direct quote, paraphrase, or faithful representation of a specific source. This is about **source fidelity**, not numerical accuracy.

**Examples from the benchmark:**
- "Summary clearly tied to Pavel Panchekha's work or field of expertise" (slides_51, checkpoint 3)
- "Summary is engaging and highlights why the event is worth attending" (slides_51, checkpoint 3)
- "At least 2 reputable, stable sources are used" (slides_51, checkpoint 3)
- "Speaker's current affiliation is explicitly named and correct according to a reputable source" (slides_51, checkpoint 4)
- "3 distinct, verifiable professional achievements are listed" (slides_51, checkpoint 4)
- "Each link is categorized correctly i.e. a link that is a blog should be categorized as a blog" (docs_37, checkpoint 3)
- "The name of each link is relevant to the webpage it opens." (docs_37, checkpoint 3)
- "Sources cited correspond to specific claims in the topic summary" (slides_51, checkpoint 6)

**Implementation signals in evaluator.py:**
- LLM calls that judge relevance, accuracy of claims, or source quality
- `evaluate_with_llm()` for content verification
- Web fetching to verify content against source pages
- `fetch_page_text_content()` used to validate claims against source material

### 6. `information_retrieval`
The step verifies that data values are correct by comparing against gold/ground-truth data. This is about **data accuracy** -- the numbers, names, or facts match expected values.

**Examples from the benchmark:**
- "Each of the 10 stocks matches a top 10 highest market cap stock in tech from the end of Q2 2023" (sheets_6, checkpoint 2)
- "The past price of each stock is correct." (sheets_6, checkpoint 2)
- "The current price of each stock is correct." (sheets_6, checkpoint 2)
- "Carbohydrates values match gold labels within tolerance (7 pts)." (sheets_2, checkpoint 6)
- "Raw Cashews link valid and matches ingredient (via HTML parsing)." (sheets_2, checkpoint 5)
- "The slide number list for each reference matches the gold list of slides" (docs_37, checkpoint 3)
- "The bar chart correctly shows the total value of each stock as a percentage of the overall portfolio." (sheets_6, checkpoint 4)
- "All reference links under each category of each lecture are present." (docs_37, checkpoint 3)
- "Each lecture has an associated title in 'Month/Day: Lecture title' format." (docs_37, checkpoint 1)
- "Event name reads 'Foundations of Web Browser Internals'" (slides_51, checkpoint 1)
- "Subheader contains date (07/27/2026)" (slides_51, checkpoint 1)

**Implementation signals in evaluator.py:**
- Comparisons against gold CSV/JSON data files
- `numerical_match_with_error()`, `text_exact_match_contained()`, `text_fuzzy_match_contained()`
- `keywords_match_robust()`, `keyword_exact_match()`
- Gold label lookups and tolerance checks
- `validate_chart_values_match()`, `validate_chart_categories_match()`

### 7. `web_visit_check`
The step checks whether specific URLs appear in the agent's browsing history. **Important:** Even if the implementation fetches page content and uses an LLM to verify that the visited page contains relevant information, the step is still a `web_visit_check` as long as its purpose is verifying browsing history. The LLM in this case is just a smarter URL-matching mechanism -- the step's intent is "did the agent visit the right page?", not "is the content correct?".

**Examples from the benchmark:**
- "The browsing history contains a visit to Tom Hanks' Wikipedia page." (slides_30, checkpoint 4)
- "The website trace contains a URL that has information about the top 10 highest market cap stocks in tech from the end of Q2 2023." (sheets_6, checkpoint 3)
- "Recipe URL (rainbowplantlife.com) visited (1 pt)." (sheets_2, checkpoint 8)
- "USDA database URL visited for Raw Cashews (1 pt)." (sheets_2, checkpoint 8)

**Implementation signals in evaluator.py:**
- Browsing history list iteration
- URL pattern matching against history entries
- `keywords_match_robust()` applied to URL strings
- LLM calls that fetch and analyze page content to verify the visited URL is relevant (this is still `web_visit_check`, not `content_check`)

---

## LLM/VLM Primary Flag Rules

Set `llm_vlm_primary` to `true` ONLY when an LLM or VLM is the **primary** evaluation mechanism for that step -- meaning the core pass/fail decision depends on the model's judgment, not on deterministic logic.

### `true` -- LLM/VLM is primary
- `binary_judge_image()` -- VLM makes the yes/no decision
- `evaluate_with_llm()` -- LLM judges content quality/relevance
- Direct `model(messages)` calls where the model's response determines success
- `verify_image_in_region()` when VLM is used for image identification (not just position)
- `match_image_tiered()` -- uses VLM as the final tier, but the step's purpose is image verification so VLM is the intended primary method

### `false` -- LLM/VLM is fallback or not used
- `keywords_match_robust()` -- keyword matching first, LLM only as fallback
- `match_columns()` -- keyword matching first, LLM fallback for fuzzy column name matching
- `text_exact_match_contained()`, `text_fuzzy_match_contained()` -- deterministic string operations
- `keyword_exact_match()` -- pure string matching
- `numerical_match_with_error()` -- arithmetic comparison
- `image_exact_match()`, `perceptual_hash_match()` -- deterministic image comparison (no LLM)
- Color, font, position checks -- all deterministic API data inspection
- Browsing history URL checks -- string matching

### How to determine
1. Find the eval step in `evaluator.py` -- locate the code that adds the `EvaluationStep` or calls `checkpoint.add_step()`.
2. Trace what function(s) determine the `success` value for that step.
3. If the function is deterministic (string match, numeric comparison, pixel comparison, API property check), set `false`.
4. If the function calls an LLM/VLM and that call's output directly determines pass/fail, set `true`.
5. If there is a tiered approach (try deterministic first, fall back to LLM), set `false` -- the primary mechanism is deterministic.
6. When in doubt, check the task-level `utils.py` to trace helper function implementations.

---

## Handling Ambiguous Steps

Some eval steps may not fit cleanly into a single category. Common ambiguities:

- **Structural vs. Information Retrieval**: "Column X exists" is structural. "Column X contains the right values" is information retrieval. If a step checks both existence AND correctness, categorize by its primary purpose.
- **Formatting vs. Spatial**: "Text is centered" is formatting (alignment property). "Text box is at the top of the page" is spatial (position on canvas). "Text is left-aligned" is formatting.
- **Content vs. Information Retrieval**: "Summary is tied to the speaker's work" is content (source fidelity). "The date is 07/27/2026" is information retrieval (exact value match).
- **Visual vs. Spatial**: "Image is present on the slide" is visual (identifying the image). "Image is in the top right" is spatial (position). If a step checks both ("Image X is in the top right"), categorize as spatial if position is the primary thing being validated, or visual if image identity is the primary thing.

When a step is genuinely ambiguous, add it to the `flagged_steps` array with:
- The checkpoint name
- The step name
- The full step description from checkpoints.md
- Your reason for flagging (e.g., "Could be structural or information_retrieval -- checks both column existence and value correctness in the same step")

---

## Output Format

For each instance, write a JSON file to `output/taxonomy/<task_name>__<instance>.json`:

```json
{
  "task_name": "sheets_6_investmenttracker",
  "instance": "instance_1",
  "eval_steps": [
    {
      "checkpoint": "Checkpoint 1",
      "step_name": "Stock Name Column",
      "category": "structural_check",
      "llm_vlm_primary": false
    },
    {
      "checkpoint": "Checkpoint 2",
      "step_name": "Stocks Match Gold Data",
      "category": "information_retrieval",
      "llm_vlm_primary": true
    }
  ],
  "flagged_steps": [
    {
      "checkpoint": "Checkpoint 3",
      "step_name": "Some Ambiguous Step",
      "description": "Full step description from checkpoints.md",
      "reason": "Could be content_check or information_retrieval -- verifies data but also checks source"
    }
  ]
}
```

**Field details:**
- `checkpoint`: The checkpoint header (e.g., "Checkpoint 1", "Checkpoint 2")
- `step_name`: A short name for the step. Use the step description from checkpoints.md, truncated to a concise label if it's long.
- `category`: Exactly one of: `visual_check`, `structural_check`, `formatting_check`, `spatial_check`, `content_check`, `information_retrieval`, `web_visit_check`
- `llm_vlm_primary`: `true` or `false` per the rules above
- `flagged_steps`: Array of steps you couldn't cleanly categorize. Can be empty.

---

## Process

1. **Work through each task one at a time.** For each task, process all 5 instances.

2. **Treat each instance independently.** Do not assume that instances of the same task have identical checkpoints -- some tasks have slight variations across instances (different numbers of items, different point values, different step counts).

3. **For each instance:**
   a. Read `checkpoints.md` to get the list of eval steps and their descriptions.
   b. Read `evaluator.py` to understand how each step is implemented.
   c. Read the task-level `utils.py` (if it exists) to trace any helper functions.
   d. For each eval step, assign a category and LLM/VLM flag.
   e. Write the JSON output file.
   f. **Verify step count**: The number of entries in your `eval_steps` array MUST exactly match the number of bullet points under `### Outcome Evaluation` sections in `checkpoints.md`. Count only bullets under Outcome Evaluation headers -- do NOT count bullets under `### Eval Template(s)` or other sections. If your count doesn't match, re-read the checkpoints.md and find the steps you missed or double-counted.

   **Warning:** Different instances of the same task can have different numbers of eval steps (e.g., one instance may have 3 contact fields while another has 4). Do not copy step counts from one instance to another -- always count from the actual checkpoints.md.

4. **After processing all instances of a task**, briefly report:
   - How many total steps were categorized
   - How many steps were flagged as ambiguous
   - Any patterns or issues you noticed

5. **After processing ALL tasks**, provide a summary:
   - Total steps categorized across all 110 instances
   - Total flagged steps
   - List all flagged steps grouped by task, so I can review them and tell you how to resolve them

---

## Task Directory Structure

All tasks are under: `src/browsergym/knows/eval/tasks/`

```
<task_name>/
  utils.py              (shared helpers -- may not exist)
  instance_1/
    checkpoints.md
    evaluator.py
    task.md
    id.txt
    data/               (gold data files)
  instance_2/
    ...
  instance_3/
    ...
  instance_4/
    ...
  instance_5/
    ...
```

The 22 tasks are:
- docs_1_formal_letter
- docs_5_influential_papers
- docs_11_personal_recipe_ocr
- docs_31_education_lesson_plan
- docs_37_reference_list
- sheets_2_personal_recipe_foodcomposition
- sheets_6_investmenttracker
- sheets_7_running_analysis
- sheets_10_paper_sorting
- sheets_25_skitourplan
- sheets_28_personal_travel_planner
- sheets_38_apartment_finder
- sheets_45_Personal_WeddingPlanner_weddingcolorpallette
- sheets_55_Movie_Recommendation
- slides_17_removeimagesaddplaceholders
- slides_20_Illustrated_Book_Report
- slides_26_basic_educational_slide_deck
- slides_29_buy_car_pres
- slides_30_Work_Wikipedia_Photos
- slides_39_Personal_Lookbook_PaintColors
- slides_42_personal_none_product_comparison
- slides_51_event_announcement_poster

Output directory: `output/taxonomy/`

\end{tcblisting}
\twocolumn

%% file: acl_latex.bbl
\begin{thebibliography}{43}
\providecommand{\natexlab}[1]{#1}

\bibitem[{Akkil et~al.(2026)Akkil, Allaham, Raj, Abuelsaad, and
  Kokku}]{akkil2026emergencewebvoyagerconsistenttransparent}
Deepak Akkil, Mowafak Allaham, Amal Raj, Tamer Abuelsaad, and Ravi Kokku. 2026.
\newblock \href {https://arxiv.org/abs/2603.29020} {Emergence webvoyager:
  Toward consistent and transparent evaluation of (web) agents in the wild}.
\newblock \emph{Preprint}, arXiv:2603.29020.

\bibitem[{Anthropic(2026)}]{claude-opus-4.7-system-card}
Anthropic. 2026.
\newblock \href
  {https://cdn.sanity.io/files/4zrzovbb/website/037f06850df7fbe871e206dad004c3db5fd50340.pdf}
  {{System Card: Claude Opus 4.7}}.

\bibitem[{Arora et~al.(2025)Arora, Wei, Hicks, Bowman, Quiñonero-Candela,
  Tsimpourlas, Sharman, Shah, Vallone, Beutel, Heidecke, and
  Singhal}]{arora2025healthbenchevaluatinglargelanguage}
Rahul~K. Arora, Jason Wei, Rebecca~Soskin Hicks, Preston Bowman, Joaquin
  Quiñonero-Candela, Foivos Tsimpourlas, Michael Sharman, Meghan Shah, Andrea
  Vallone, Alex Beutel, Johannes Heidecke, and Karan Singhal. 2025.
\newblock \href {https://arxiv.org/abs/2505.08775} {Healthbench: Evaluating
  large language models towards improved human health}.
\newblock \emph{Preprint}, arXiv:2505.08775.

\bibitem[{Caldwell et~al.(2025)Caldwell, Harley, Kouremetis, Abruzzo, and
  Pearce}]{caldwell2025pentestjudgejudgingagentbehavior}
Shane Caldwell, Max Harley, Michael Kouremetis, Vincent Abruzzo, and Will
  Pearce. 2025.
\newblock \href {https://arxiv.org/abs/2508.02921} {Pentestjudge: Judging agent
  behavior against operational requirements}.
\newblock \emph{Preprint}, arXiv:2508.02921.

\bibitem[{Chen et~al.(2024)Chen, Chen, Zhang, Wang, Liu, Zhou, Zhang, Wan,
  Zhou, and Sun}]{chen2024mllm}
Dongping Chen, Ruoxi Chen, Shilin Zhang, Yaochen Wang, Yinuo Liu, Huichi Zhou,
  Qihui Zhang, Yao Wan, Pan Zhou, and Lichao Sun. 2024.
\newblock \href {https://openreview.net/forum?id=dbFEFHAD79}
  {{MLLM}-as-a-judge: Assessing multimodal {LLM}-as-a-judge with
  vision-language benchmark}.
\newblock In \emph{Forty-first International Conference on Machine Learning}.

\bibitem[{Chen et~al.(2026)Chen, Zhu, Li, Wang, Yang, and
  Guo}]{chen2026presentbench}
Xin-Sheng Chen, Jiayu Zhu, Pei-lin Li, Hanzheng Wang, Shuojin Yang, and
  Meng-Hao Guo. 2026.
\newblock \href {https://arxiv.org/abs/2603.07244} {Presentbench: A
  fine-grained rubric-based benchmark for slide generation}.
\newblock \emph{arXiv preprint arXiv:2603.07244}.

\bibitem[{Cho et~al.(2023)Cho, Zala, and Bansal}]{10.5555/3666122.3666387}
Jaemin Cho, Abhay Zala, and Mohit Bansal. 2023.
\newblock \href {https://arxiv.org/abs/2305.15328} {Visual programming for
  text-to-image generation and evaluation}.
\newblock In \emph{Proceedings of the 37th International Conference on Neural
  Information Processing Systems}, NIPS '23, Red Hook, NY, USA. Curran
  Associates Inc.

\bibitem[{de~Chezelles et~al.(2025)de~Chezelles, Gasse, Lacoste, Caccia,
  Drouin, Boisvert, Thakkar, Marty, Assouel, Shayegan, Jang, L{\`u}, Yoran,
  Kong, Xu, Reddy, Neubig, Cappart, Salakhutdinov, and
  Chapados}]{chezelles2025the}
Thibault Le~Sellier de~Chezelles, Maxime Gasse, Alexandre Lacoste, Massimo
  Caccia, Alexandre Drouin, L{\'e}o Boisvert, Megh Thakkar, Tom Marty, Rim
  Assouel, Sahar~Omidi Shayegan, Lawrence~Keunho Jang, Xing~Han L{\`u}, Ori
  Yoran, Dehan Kong, Frank~F. Xu, Siva Reddy, Graham Neubig, Quentin Cappart,
  Russ Salakhutdinov, and Nicolas Chapados. 2025.
\newblock \href {https://openreview.net/forum?id=5298fKGmv3} {The browsergym
  ecosystem for web agent research}.
\newblock \emph{Transactions on Machine Learning Research}.
\newblock Expert Certification.

\bibitem[{{DeepSeek-AI}(2026)}]{deepseek2026v4}
{DeepSeek-AI}. 2026.
\newblock {DeepSeek-V4}: Towards highly efficient million-token context
  intelligence.
\newblock Technical report, DeepSeek.
\newblock Preview Release. Technical report available at
  \url{https://huggingface.co/deepseek-ai/DeepSeek-V4-Pro/blob/main/DeepSeek_V4.pdf}.

\bibitem[{Deng et~al.(2023)Deng, Gu, Zheng, Chen, Stevens, Wang, Sun, and
  Su}]{deng2023mind2webgeneralistagentweb}
Xiang Deng, Yu~Gu, Boyuan Zheng, Shijie Chen, Samuel Stevens, Boshi Wang, Huan
  Sun, and Yu~Su. 2023.
\newblock \href {https://openreview.net/forum?id=kiYqbO3wqw} {Mind2web: Towards
  a generalist agent for the web}.
\newblock In \emph{Thirty-seventh Conference on Neural Information Processing
  Systems Datasets and Benchmarks Track}.

\bibitem[{Drouin et~al.(2024)Drouin, Gasse, Caccia, Laradji, Del~Verme, Marty,
  Vazquez, Chapados, and Lacoste}]{workarena2024}
Alexandre Drouin, Maxime Gasse, Massimo Caccia, Issam~H. Laradji, Manuel
  Del~Verme, Tom Marty, David Vazquez, Nicolas Chapados, and Alexandre Lacoste.
  2024.
\newblock \href {https://proceedings.mlr.press/v235/drouin24a.html}
  {{W}ork{A}rena: How capable are web agents at solving common knowledge work
  tasks?}
\newblock In \emph{Proceedings of the 41st International Conference on Machine
  Learning}, volume 235 of \emph{Proceedings of Machine Learning Research},
  pages 11642--11662. PMLR.

\bibitem[{Du et~al.(2026)Du, Xu, Zhu, Zhang, Wang, and
  Mao}]{du2026deepresearch}
Mingxuan Du, Benfeng Xu, Chiwei Zhu, Licheng Zhang, Xiaorui Wang, and Zhendong
  Mao. 2026.
\newblock \href {https://openreview.net/forum?id=hQ0K2Hhq7H} {Deepresearch
  bench: A comprehensive benchmark for deep research agents}.
\newblock In \emph{The Fourteenth International Conference on Learning
  Representations}.

\bibitem[{Gilardi et~al.(2023)Gilardi, Alizadeh, and Kubli}]{Gilardi_2023}
Fabrizio Gilardi, Meysam Alizadeh, and Maël Kubli. 2023.
\newblock \href {https://doi.org/10.1073/pnas.2305016120} {Chatgpt outperforms
  crowd workers for text-annotation tasks}.
\newblock \emph{Proceedings of the National Academy of Sciences}, 120(30).

\bibitem[{Gou et~al.(2026)Gou, Huang, Ning, Gu, Lin, Qi, Kopanev, Yu,
  Gutierrez, Shu, Song, Wu, Chen, Moussa, ZHANG, Xie, Li, Xue, Liao, Zhang,
  Zheng, Cai, Rozgic, Ziyadi, Sun, and Su}]{gou2025mind2web}
Boyu Gou, Zanming Huang, Yuting Ning, Yu~Gu, Michael Lin, Weijian Qi, Andrei
  Kopanev, Botao Yu, Bernal~Jimenez Gutierrez, Yiheng Shu, Chan~Hee Song,
  Jiaman Wu, Shijie Chen, Hanane~Nour Moussa, TIANSHU ZHANG, Jian Xie, Yifei
  Li, Tianci Xue, Zeyi Liao, and 7 others. 2026.
\newblock \href {https://openreview.net/forum?id=AUaW6DS9si} {Mind2web 2:
  Evaluating agentic search with agent-as-a-judge}.
\newblock In \emph{The Thirty-ninth Annual Conference on Neural Information
  Processing Systems Datasets and Benchmarks Track}.

\bibitem[{Han et~al.(2026)Han, Kim, Lee, Lee, Park, Song, Choi, Lee, and
  Lee}]{han2026deerbenchmarkevaluatingdeep}
Janghoon Han, Heegyu Kim, Changho Lee, Dahm Lee, Min~Hyung Park, Hosung Song,
  Stanley~Jungkyu Choi, Moontae Lee, and Honglak Lee. 2026.
\newblock \href {https://arxiv.org/abs/2512.17776} {Deer: A benchmark for
  evaluating deep research agents on expert report generation}.
\newblock \emph{Preprint}, arXiv:2512.17776.

\bibitem[{Jia et~al.(2026)Jia, Liao, Zhang, Xu, Xie, Jiang, Yan, Liu, Ye, and
  Huang}]{jia2026osworldmcp}
Hongrui Jia, Jitong Liao, Xi~Zhang, Haiyang Xu, Tianbao Xie, Chaoya Jiang, Ming
  Yan, Si~Liu, Wei Ye, and Fei Huang. 2026.
\newblock \href {https://openreview.net/forum?id=rceD6wwt4B} {{OSW}orld-{MCP}:
  Benchmarking {MCP} tool invocation in computer-use agents}.
\newblock In \emph{The Fourteenth International Conference on Learning
  Representations}.

\bibitem[{Kapoor et~al.(2026)Kapoor, Stroebl, Kirgis, Nadgir, Siegel, Wei, Xue,
  Chen, Chen, Utpala, Ndzomga, Oruganty, Luskin, Liu, Yu, Arora, Hahm, Trivedi,
  Sun, Lee, Jin, Mai, Zhou, Zhu, Bommasani, Kang, Song, Henderson, Su, Liang,
  and Narayanan}]{kapoor2026holistic}
Sayash Kapoor, Benedikt Stroebl, Peter Kirgis, Nitya Nadgir, Zachary~S Siegel,
  Boyi Wei, Tianci Xue, Ziru Chen, Felix Chen, Saiteja Utpala, Franck Ndzomga,
  Dheeraj Oruganty, Sophie Luskin, Kangheng Liu, Botao Yu, Amit Arora, Dongyoon
  Hahm, Harsh Trivedi, Huan Sun, and 12 others. 2026.
\newblock \href {https://openreview.net/forum?id=vUaY1t64ZZ} {Holistic agent
  leaderboard: The missing infrastructure for {AI} agent evaluation}.
\newblock In \emph{The Fourteenth International Conference on Learning
  Representations}.

\bibitem[{Khalifa et~al.(2026)Khalifa, Logeswaran, Kim, Sohn, Zhang, Lee, Peng,
  Wang, and Lee}]{khalifa2026gamingjudgeunfaithfulchainofthought}
Muhammad Khalifa, Lajanugen Logeswaran, Jaekyeom Kim, Sungryull Sohn, Yunxiang
  Zhang, Moontae Lee, Hao Peng, Lu~Wang, and Honglak Lee. 2026.
\newblock \href {https://arxiv.org/abs/2601.14691} {Gaming the judge:
  Unfaithful chain-of-thought can undermine agent evaluation}.
\newblock \emph{Preprint}, arXiv:2601.14691.

\bibitem[{Koh et~al.(2024)Koh, Lo, Jang, Duvvur, Lim, Huang, Neubig, Zhou,
  Salakhutdinov, and Fried}]{koh2024visualwebarenaevaluatingmultimodalagents}
Jing~Yu Koh, Robert Lo, Lawrence Jang, Vikram Duvvur, Ming Lim, Po-Yu Huang,
  Graham Neubig, Shuyan Zhou, Russ Salakhutdinov, and Daniel Fried. 2024.
\newblock \href {https://doi.org/10.18653/v1/2024.acl-long.50}
  {{V}isual{W}eb{A}rena: Evaluating multimodal agents on realistic visual web
  tasks}.
\newblock In \emph{Proceedings of the 62nd Annual Meeting of the Association
  for Computational Linguistics (Volume 1: Long Papers)}, pages 881--905,
  Bangkok, Thailand. Association for Computational Linguistics.

\bibitem[{Li et~al.(2026)Li, Choe, Liu, Chen, Tao, You, Chen, Di, Sun, Zheng
  et~al.}]{li2026clawsbench}
Xiangyi Li, Kyoung~Whan Choe, Yimin Liu, Xiaokun Chen, Chujun Tao, Bingran You,
  Wenbo Chen, Zonglin Di, Jiankai Sun, Shenghan Zheng, and 1 others. 2026.
\newblock \href {https://arxiv.org/abs/2604.05172} {Clawsbench: Evaluating
  capability and safety of llm productivity agents in simulated workspaces}.
\newblock \emph{arXiv preprint arXiv:2604.05172}.

\bibitem[{Lowe(1999)}]{lowe1999object}
David~G Lowe. 1999.
\newblock \href {https://ieeexplore.ieee.org/abstract/document/790410} {Object
  recognition from local scale-invariant features}.
\newblock In \emph{Proceedings of the seventh IEEE international conference on
  computer vision}, volume~2, pages 1150--1157. Ieee.

\bibitem[{L\`{u} et~al.(2024)L\`{u}, Kasner, and
  Reddy}]{lu2024weblinxrealworldwebsitenavigation}
Xing~Han L\`{u}, Zden\v{e}k Kasner, and Siva Reddy. 2024.
\newblock \href {https://arxiv.org/abs/2402.05930} {Weblinx: real-world website
  navigation with multi-turn dialogue}.
\newblock In \emph{Proceedings of the 41st International Conference on Machine
  Learning}, ICML'24. JMLR.org.

\bibitem[{L{\`u} et~al.(2025)L{\`u}, Kazemnejad, Meade, Patel, Shin, Zambrano,
  Stanczak, Shaw, Pal, and Reddy}]{lu2025agentrewardbench}
Xing~Han L{\`u}, Amirhossein Kazemnejad, Nicholas Meade, Arkil Patel, Dongchan
  Shin, Alejandra Zambrano, Karolina Stanczak, Peter Shaw, Christopher Pal, and
  Siva Reddy. 2025.
\newblock \href {https://openreview.net/forum?id=fQcUZMPIvu} {Agentrewardbench:
  Evaluating automatic evaluations of web agent trajectories}.
\newblock In \emph{Second Conference on Language Modeling}.

\bibitem[{Ma et~al.(2024)Ma, Zhang, Zhang, Yu, Zhang, Zhang, Luo, Wang, and
  Tang}]{ma2024spreadsheetbench}
Zeyao Ma, Bohan Zhang, Jing Zhang, Jifan Yu, Xiaokang Zhang, Xiaohan Zhang,
  Sijia Luo, Xi~Wang, and Jie Tang. 2024.
\newblock \href {https://openreview.net/forum?id=KYxzmRLF6i} {Spreadsheetbench:
  Towards challenging real world spreadsheet manipulation}.
\newblock In \emph{The Thirty-eight Conference on Neural Information Processing
  Systems Datasets and Benchmarks Track}.

\bibitem[{{OpenAI}(2025)}]{openai2025atlas}
{OpenAI}. 2025.
\newblock Introducing {ChatGPT Atlas}.
\newblock \url{https://openai.com/index/introducing-chatgpt-atlas/}.
\newblock Accessed: 2026-05-25.

\bibitem[{OpenAI(2026)}]{gpt-5.5-system-card}
OpenAI. 2026.
\newblock \href {https://deploymentsafety.openai.com/gpt-5-5/gpt-5-5.pdf}
  {{GPT}-5.5 {S}ystem {Card}}.

\bibitem[{Pan et~al.(2024)Pan, Kong, Zhou, Cui, Leng, Jiang, Liu, Shang, Zhou,
  Wu et~al.}]{pan2024webcanvas}
Yichen Pan, Dehan Kong, Sida Zhou, Cheng Cui, Yifei Leng, Bing Jiang, Hangyu
  Liu, Yanyi Shang, Shuyan Zhou, Tongshuang Wu, and 1 others. 2024.
\newblock \href {https://arxiv.org/abs/2406.12373} {Webcanvas: Benchmarking web
  agents in online environments}.
\newblock \emph{arXiv preprint arXiv:2406.12373}.

\bibitem[{Sharma et~al.(2026)Sharma, Zhang, Bandi, Wang, Aich, Nghiem, Rabbani,
  Htet, Jang, Basu, Balwani, Peskoff, Ayestaran, Hendryx, Kenstler, and
  Liu}]{sharma2025researchrubrics}
Manasi Sharma, Chen Bo~Calvin Zhang, Chaithanya Bandi, Clinton Wang, Ankit
  Aich, Huy Nghiem, Tahseen Rabbani, Ye~Htet, Brian Jang, Sumana Basu,
  Aishwarya Balwani, Denis Peskoff, Marcos Ayestaran, Sean~M. Hendryx, Brad
  Kenstler, and Bing Liu. 2026.
\newblock \href {https://openreview.net/forum?id=ErnvfmSX0P} {Researchrubrics:
  A benchmark of prompts and rubrics for evaluating deep research agents}.
\newblock In \emph{The Fourteenth International Conference on Learning
  Representations}.

\bibitem[{Shen et~al.(2026)Shen, Qiu, Whitehouse, Alazraki, Goel, Barbieri,
  Willi, Mathur, and Leontiadis}]{shen2026rethinkingrubricgenerationimproving}
William~F. Shen, Xinchi Qiu, Chenxi Whitehouse, Lisa Alazraki, Shashwat Goel,
  Francesco Barbieri, Timon Willi, Akhil Mathur, and Ilias Leontiadis. 2026.
\newblock \href {https://arxiv.org/abs/2602.05125} {Rethinking rubric
  generation for improving llm judge and reward modeling for open-ended tasks}.
\newblock \emph{Preprint}, arXiv:2602.05125.

\bibitem[{Song et~al.(2025)Song, Thai, Pham, Chang, Nadaf, and
  Iyyer}]{song2025bearcubs}
Yixiao Song, Katherine Thai, Chau~Minh Pham, Yapei Chang, Mazin Nadaf, and
  Mohit Iyyer. 2025.
\newblock \href {https://openreview.net/forum?id=0JzWiigkUy} {{BEARCUBS}: A
  benchmark for computer-using web agents}.
\newblock In \emph{Second Conference on Language Modeling}.

\bibitem[{Sun and Chen(2025)}]{sun2025webarxivevaluatingmultimodalagents}
Zihao Sun and Ling Chen. 2025.
\newblock \href {https://arxiv.org/abs/2507.00938} {Webarxiv: Evaluating
  multimodal agents on time-invariant arxiv tasks}.
\newblock \emph{Preprint}, arXiv:2507.00938.

\bibitem[{Wang et~al.(2026)Wang, He, Chen, Yehudai, Liu, Ying, Shmueli-Scheuer,
  and Cohan}]{wang2026timereflecttrustllm}
Leyao Wang, Yanan He, Peng Chen, Asaf Yehudai, Yixin Liu, Rex Ying, Michal
  Shmueli-Scheuer, and Arman Cohan. 2026.
\newblock \href {https://arxiv.org/abs/2605.19196} {Time to reflect: Can we
  trust llm judges for evidence-based research agents?}
\newblock \emph{Preprint}, arXiv:2605.19196.

\bibitem[{Wang et~al.(2025)Wang, Han, Diaz, Xu, R{\"u}hle, and
  Rajmohan}]{wang2025odysseybench}
Weixuan Wang, Dongge Han, Daniel~Madrigal Diaz, Jin Xu, Victor R{\"u}hle, and
  Saravan Rajmohan. 2025.
\newblock \href {https://arxiv.org/abs/2508.09124} {Odysseybench: Evaluating
  llm agents on long-horizon complex office application workflows}.
\newblock \emph{arXiv preprint arXiv:2508.09124}.

\bibitem[{Wang et~al.(2024)Wang, Cui, Zhong, Zhang, Yin, Lin, and
  Shang}]{wang2024officebench}
Zilong Wang, Yuedong Cui, Li~Zhong, Zimin Zhang, Da~Yin, Bill~Yuchen Lin, and
  Jingbo Shang. 2024.
\newblock Officebench: Benchmarking language agents across multiple
  applications for office automation.
\newblock \emph{arXiv preprint arXiv:2407.19056}.

\bibitem[{Wei et~al.(2025)Wei, Sun, Papay, McKinney, Han, Fulford, Chung,
  Passos, Fedus, and Glaese}]{wei2025browsecompsimplechallengingbenchmark}
Jason Wei, Zhiqing Sun, Spencer Papay, Scott McKinney, Jeffrey Han, Isa
  Fulford, Hyung~Won Chung, Alex~Tachard Passos, William Fedus, and Amelia
  Glaese. 2025.
\newblock \href {https://arxiv.org/abs/2504.12516} {Browsecomp: A simple yet
  challenging benchmark for browsing agents}.
\newblock \emph{Preprint}, arXiv:2504.12516.

\bibitem[{Xie et~al.(2024)Xie, Zhang, Chen, Li, Zhao, Cao, Hua, Cheng, Shin,
  Lei, Liu, Xu, Zhou, Savarese, Xiong, Zhong, and
  Yu}]{xie2024osworldbenchmarkingmultimodalagents}
Tianbao Xie, Danyang Zhang, Jixuan Chen, Xiaochuan Li, Siheng Zhao, Ruisheng
  Cao, Toh~Jing Hua, Zhoujun Cheng, Dongchan Shin, Fangyu Lei, Yitao Liu,
  Yiheng Xu, Shuyan Zhou, Silvio Savarese, Caiming Xiong, Victor Zhong, and Tao
  Yu. 2024.
\newblock \href {https://openreview.net/forum?id=tN61DTr4Ed} {{OSW}orld:
  Benchmarking multimodal agents for open-ended tasks in real computer
  environments}.
\newblock In \emph{The Thirty-eight Conference on Neural Information Processing
  Systems Datasets and Benchmarks Track}.

\bibitem[{Xue et~al.(2025)Xue, Qi, Shi, Song, Gou, Song, Sun, and
  Su}]{xue2025an}
Tianci Xue, Weijian Qi, Tianneng Shi, Chan~Hee Song, Boyu Gou, Dawn Song, Huan
  Sun, and Yu~Su. 2025.
\newblock \href {https://openreview.net/forum?id=6jZi4HSs6o} {An illusion of
  progress? assessing the current state of web agents}.
\newblock In \emph{Second Conference on Language Modeling}.

\bibitem[{Yang et~al.(2025)Yang, Yonack, Zyskowski, Yarats, Ho, and
  Ma}]{yang2025adoptionusageaiagents}
Jeremy Yang, Noah Yonack, Kate Zyskowski, Denis Yarats, Johnny Ho, and Jerry
  Ma. 2025.
\newblock \href {https://arxiv.org/abs/2512.07828} {The adoption and usage of
  ai agents: Early evidence from perplexity}.
\newblock \emph{Preprint}, arXiv:2512.07828.

\bibitem[{Yao et~al.(2022)Yao, Chen, Yang, and Narasimhan}]{yao2022webshop}
Shunyu Yao, Howard Chen, John Yang, and Karthik Narasimhan. 2022.
\newblock \href {https://openreview.net/forum?id=R9KnuFlvnU} {Webshop: Towards
  scalable real-world web interaction with grounded language agents}.
\newblock \emph{Advances in Neural Information Processing Systems},
  35:20744--20757.

\bibitem[{Yehudai et~al.(2026)Yehudai, Eden, Li, Uziel, Zhao, Bar-Haim, Cohan,
  and Shmueli-Scheuer}]{yehudai2026survey}
Asaf Yehudai, Lilach Eden, Alan Li, Guy Uziel, Yilun Zhao, Roy Bar-Haim, Arman
  Cohan, and Michal Shmueli-Scheuer. 2026.
\newblock \href {https://arxiv.org/abs/2503.16416} {Survey on evaluation of
  llm-based agents}.
\newblock \emph{Findings of the Association for Computational Linguistics: ACL
  2026}.

\bibitem[{Yoran et~al.(2024)Yoran, Amouyal, Malaviya, Bogin, Press, and
  Berant}]{yoran-etal-2024-assistantbench}
Ori Yoran, Samuel~Joseph Amouyal, Chaitanya Malaviya, Ben Bogin, Ofir Press,
  and Jonathan Berant. 2024.
\newblock \href {https://doi.org/10.18653/v1/2024.emnlp-main.505}
  {{A}ssistant{B}ench: Can web agents solve realistic and time-consuming
  tasks?}
\newblock In \emph{Proceedings of the 2024 Conference on Empirical Methods in
  Natural Language Processing}, pages 8938--8968, Miami, Florida, USA.
  Association for Computational Linguistics.

\bibitem[{Zheng et~al.(2023)Zheng, Chiang, Sheng, Zhuang, Wu, Zhuang, Lin, Li,
  Li, Xing, Zhang, Gonzalez, and Stoica}]{zheng2023judging}
Lianmin Zheng, Wei-Lin Chiang, Ying Sheng, Siyuan Zhuang, Zhanghao Wu, Yonghao
  Zhuang, Zi~Lin, Zhuohan Li, Dacheng Li, Eric Xing, Hao Zhang, Joseph~E.
  Gonzalez, and Ion Stoica. 2023.
\newblock \href {https://openreview.net/forum?id=uccHPGDlao} {Judging
  {LLM}-as-a-judge with {MT}-bench and chatbot arena}.
\newblock In \emph{Thirty-seventh Conference on Neural Information Processing
  Systems Datasets and Benchmarks Track}.

\bibitem[{Zhou et~al.(2024)Zhou, Xu, Zhu, Zhou, Lo, Sridhar, Cheng, Ou, Bisk,
  Fried, Alon, and Neubig}]{zhou2024webarena}
Shuyan Zhou, Frank~F. Xu, Hao Zhu, Xuhui Zhou, Robert Lo, Abishek Sridhar,
  Xianyi Cheng, Tianyue Ou, Yonatan Bisk, Daniel Fried, Uri Alon, and Graham
  Neubig. 2024.
\newblock \href {https://openreview.net/forum?id=oKn9c6ytLx} {Webarena: A
  realistic web environment for building autonomous agents}.
\newblock In \emph{The Twelfth International Conference on Learning
  Representations}.

\end{thebibliography}
